\documentclass[lettersize,journal]{IEEEtran}

\usepackage{amsmath,amsfonts}
\usepackage{algorithmic}
\usepackage{algorithm}
\usepackage{array}
\usepackage[caption=false,font=normalsize,labelfont=sf,textfont=sf]{subfig}
\usepackage{textcomp}
\usepackage{stfloats}
\usepackage{url}
\usepackage{verbatim}
\usepackage{graphicx}
\usepackage{cite}
\usepackage{multirow}

\usepackage{caption}
\usepackage{siunitx}

\graphicspath{{figures/}} 

\usepackage[acronym]{glossaries}
\newacronym{AHRS}{AHRS}{Attitude and Heading Reference System}
\newacronym{AI}{AI}{Artificial Intelligence}
\newacronym[plural=ADeMs,firstplural=Autonomous Agents]{ADeM}{ADeM}{Autonomous Agent}
\newacronym[plural=APIs,firstplural=Application Programming Interfaces]{API}{API}{Application Programming Interface}
\newacronym{AAW}{AAW}{Anti-Aircraft Warfare}
\newacronym{ASW}{ASW}{Anti-Submarine Warfare}
\newacronym{ATP}{ATP}{Acquisition, tracking and pointing}
\newacronym[plural=AUVs,firstplural=Autonomous Underwater Vehicles]{AUV}{AUV}{Autonomous Underwater Vehicle}
\newacronym[plural=BTs,firstplural=Behaviour Trees]{BT}{BT}{Behaviour Tree}
\newacronym{CAI}{CAI}{Cold Atom Interferometry}
\newacronym{CO}{CO}{Commanding Officer} 
\newacronym{CPU}{CPU}{Central Processing Unit}
\newacronym{CSAC}{CSAC}{Chip Scale Atomic Clock}
\newacronym{CSMA}{CSMA}{Carrier-Sense Multiple Access}
\newacronym{COTS}{COTS}{Commercial off-the-shelf}
\newacronym{c-QED}{c-QED}{cavity Quantum ElectroDynamics }
\newacronym{Dstl}{Dstl}{Defence Science and Technology Laboratory }
\newacronym{DVL}{DVL}{Doppler Velocity Log}
\newacronym{DR}{DR}{Dead-Reckoning}
\newacronym{EKF}{EKF}{Extended Kalman Filter}
\newacronym{GIS}{GIS}{Geographic Information System}
\newacronym{GNC}{GNC}{Guidance, Navigation, Control}
\newacronym{GNSS}{GNSS}{Global Navigation Satellite Systems}
\newacronym{GPS}{GPS}{Global Positioning System}
\newacronym{WASDGPS}{WASDGPS}{Wide Area Differential GPS}
\newacronym{RTK}{RTK}{Real-time kinematic}
\newacronym[plural=GUIs,firstplural=Graphical User Interfaces]{GUI}{GUI}{Graphical User Interface}
\newacronym{HOOTL}{HOOTL}{Human Out Of The Loop}
\newacronym{HOM}{HOM}{Hong–Ou–Mandel}
\newacronym{IMU}{IMU}{Inertial Measurement Unit}
\newacronym{INS}{INS}{Inertial Navigation System}
\newacronym{ISAIN}{ISAIN}{Intelligent Ship AI Network}
\newacronym{LBL}{LBL}{Long Base Line}
\newacronym{LCAT}{LCAT}{Low-Cost AUV Technology}
\newacronym{MAC}{MAC}{Medium Access Control}
\newacronym{MCM}{MCM}{Mine Counter Measures}
\newacronym{MZI}{MZI}{Mach-Zehnder Interferometer}
\newacronym[plural=MPAs,firstplural=Maritime Patrol Aircrafts]{MPA}{MPA}{Maritime Patrol Aircraft}
\newacronym{MEMS}{MEMS}{Micro-Electro-Mechanical System}
\newacronym{MOFE}{MOFE}{magneto-optic Faraday effect}
\newacronym{MOT}{magneto-optical trap}{magneto-optical trap}
\newacronym{NASA}{NASA}{National Aeronautics and Space Administration}
\newacronym{NE}{NE}{North-East}
\newacronym[plural=OEXs, firstplural=Ocean Explorers]{OEX}{OEX}{Ocean Explorer}
\newacronym{PF}{PF}{Particle Filter}
\newacronym{PRN}{PRN}{PseudoRandom Noise}
\newacronym{QCL}{QCL}{Quantum Cascade Laser}
\newacronym[plural=OAMs,firstplural=Optical Atomic Magnetometers]{OAM}{OAM}{Optical Atomic Magnetometer}
\newacronym{OWTT}{OWTT}{one-way travel time}
\newacronym{QKD}{QKD}{Quantum Key Distribution} 
\newacronym{rlsAO}{rLS.AO}{robustifying recursive Least Squares  for  Additive  Outliers}
\newacronym{RAS}{RAS}{Robotics and Autonomous Systems}
\newacronym{RF-OAM}{RF-OAM}{radio-frequency OAM}
\newacronym{ROS}{ROS}{Robot Operating System}
\newacronym{SA}{SA}{Single Access}
\newacronym{SAW}{SAW}{Surface Acoustic Wave}
\newacronym{SBL}{SBL}{Short Base Line}
\newacronym{SERF}{SERF}{Spin-Exchange Relaxation-Free}
\newacronym{SCI}{SCI}{Symmetrized Composite-pulse Interferometer}
\newacronym{SLAM}{SLAM}{Simultaneous Localisation And Mapping}
\newacronym{SLM}{SLM}{Spatial Light Modulator}
\newacronym{SNR}{SNR}{Signal To Noise Ratio}
\newacronym{SONA}{SONA}{Service-Oriented Network Architecture}
\newacronym{SPDC}{SPDC}{Spontaneous Parametric Down Conversion}
\newacronym[plural=SQUIDs,firstplural=Superconducting Quantum Interference Devices]{SQUID}{SQUID}{Superconducting Quantum Interference Device}
\newacronym{TDMA}{TDMA}{Time-Division Multiple Access}
\newacronym{TOA}{TOA}{Time of Arrival}
\newacronym{TDOA}{TDOA}{Time Difference of Arrival}
\newacronym{TTL}{TTL}{Time to Live}
\newacronym{TWTT}{TWTT}{Two-Way Travel Time}
\newacronym{USBL}{USBL}{Ultra-Short Base Line}
\newacronym{UAV}{UAV}{Unmanned Autonomous Vehicle}
\newacronym{UUV}{UUV}{Unmanned Underwater Vehicle}
\newacronym[plural=USVs,firstplural=Unmanned Surface Vehicles]{USV}{USV}{Unmanned Surface Vehicle}
\newacronym{UKF}{UKF}{Unscented Kalman Filter}
\newacronym{URI}{URI}{Uniform Resource Identifier}
\newacronym{QPS}{QPS}{Quantum Positioning System}
\newacronym{GDOP}{GDOP}{Geometric Dilution of Precision}
\newacronym{RF}{RF}{Radio Frequency}   
\newacronym{TRL}{TRL}{Technology Readiness Level}
\newacronym{NCSC}{NCSC}{National Cyber Security Centre}

\newcommand{\neswarrow}{\mathrel{\text{$\nearrow$\llap{$\swarrow$}}}}
\newcommand{\nwsearrow}{\mathrel{\text{$\nwarrow$\llap{$\searrow$}}}}

\IEEEoverridecommandlockouts
\usepackage{tikz}
\usepackage{textcomp}
\usepackage{hyperref}
\usepackage{lipsum}

\newcommand\copyrighttext{%
  \footnotesize \textcopyright 2023 
  This work has been submitted to the IEEE for possible publication. Copyright may be transferred without notice, after which this version may no longer be accessible.}
\newcommand\copyrightnotice{%
\begin{tikzpicture}[remember picture,overlay]
\node[anchor=south,yshift=10pt] at (current page.south) {\fbox{\parbox{\dimexpr\textwidth-\fboxsep-\fboxrule\relax}{\copyrighttext}}};
\end{tikzpicture}%
}

\begin{document}

\title{Current Trends and Advances in Quantum Navigation for Maritime Applications: A Comprehensive Review}


\author{Olga Sambataro, Riccardo Costanzi, Joao Alves, Andrea Caiti, Pietro Paglierani, Roberto Petroccia, Andrea Munaf\`o$^*$~\IEEEmembership{Senior Member,~IEEE,}
\thanks{O. Sambataro, R. Costanzi, A. Caiti, A. Munaf\`o (*andrea.munafo@unipi.it - corresponding author) are with the Dept. Of Information Engineering, University of Pisa, and with the Interuniversity Centre of Integrated Systems for the Marine Environment; J. Alves, P. Paglierani and R. Petroccia are with the NATO STO Centre of Maritime Research and Experimentation}
}

\markboth{Journal of \LaTeX\ Class Files,~Vol.~14, No.~8, August~2021}%
{Shell \MakeLowercase{\textit{et al.}}: A Sample Article Using IEEEtran.cls for IEEE Journals}


\maketitle
\begin{abstract}
This paper presents a comprehensive review of the current state of the art in quantum navigation systems, with a specific focus on their application in maritime navigation. 
Quantum technologies have the potential to revolutionise navigation and positioning systems due to their ability to provide highly accurate and secure information. 
The review covers the principles of quantum navigation and highlights the latest developments in quantum-enhanced sensors, atomic clocks, and quantum communication protocols. 
The paper also discusses the challenges and opportunities of using quantum technologies in maritime navigation, including the effects that the maritime environment and the specificity of marine applications can have on the performance of quantum sensors. 
Finally, the paper concludes with a discussion on the future of quantum navigation systems and their potential impact on the maritime industry. 
This review aims at providing a valuable resource for researchers and engineers interested in the development and deployment of quantum navigation systems.
\end{abstract}


\copyrightnotice 

\section{Introduction}
\label{sec:introduction}
\IEEEPARstart{C}{urrent} maritime navigation technologies are limited in their ability to produce stable and accurate location data~\cite{ZHANG2023113861}.
This is due to the fragility of satellites, which are relatively easy to hack, hijack and spoof, susceptible to adverse cosmic conditions, and very limited at high latitudes, and to the limitations of existing~\glspl{INS}~\cite{Paull:2014}.
\glspl{INS} calculate position through a series of accelerometers and gyroscopes, and are subject to drift, meaning they are only accurate for a limited time before requiring recalibration from an external (typically satellite-based) source, making them unsuitable for vehicles set to be at sea for long (e.g., days, weeks if not months).
For vehicles navigating underwater, with no access to \gls{GNSS}, inertial navigation, often coupled with \gls{DVL}, is the main means of localisation.
Alternative methods for undersea navigation rely on the deployment of additional infrastructure (e.g., \gls{LBL}~\cite{osti_6035874}) to act as an external reference, on algorithmic solutions, such as terrain-based navigation~\cite{Salavasidis} or \gls{SLAM}~\cite{7081165} to limit the navigational drift, or on period re-surfacing to obtain a satellite-based position fix before continuing the mission.
The limitations of these approaches are evident and range from the complexity of deploying infrastructure at sea, to constrained operations (e.g., they cannot be deployed in contested environments), to mission interruptions to go to the surface to obtain position data~\cite{Paull:2014}.
In some applications, moving to the surface to navigate might not be possible due to operational constraints. 
For example, in the case of very deep missions, the time to surface can be incompatible with the mission requirements, under-ice operations make access to the surface impossible, and more broadly, \gls{GNSS} denial is today recognised as a critical vulnerability by governments globally.

New emerging quantum technologies now promise the next generation of products with exciting and astounding properties that will affect the way in which we navigate and work at sea.
Quantum technologies are expected to have a profound impact on many of the world’s biggest markets. A significant part of this involves enhancing the \$ 360 billion semiconductor industry and the \$ 1.8 trillion oil and gas industry around the world~\cite{uk-roadmap}.  
While quantum communications will enable ultra-secure data transmission, quantum timing devices will coordinate the timing of data packets without overreliance on \gls{GNSS} timing signals. 
The use of quantum imaging sensors will greatly enhance low light and sub-surface imaging capabilities, leading to a substantial impact on underwater camera technology.

This paper reviews the state of the art in quantum navigation, with a focus on the opportunities specific to the maritime domain. 
Additionally, the authors’ views, according to what is available in the current literature, on what is likely to be possible in the future is discussed. 
The paper highlights the technologies that could begin the transition of research and lab-based demonstrators towards higher \gls{TRL} engineering.\\

\begin{figure*}[htp]
    \centering
    \includegraphics[width=\textwidth]{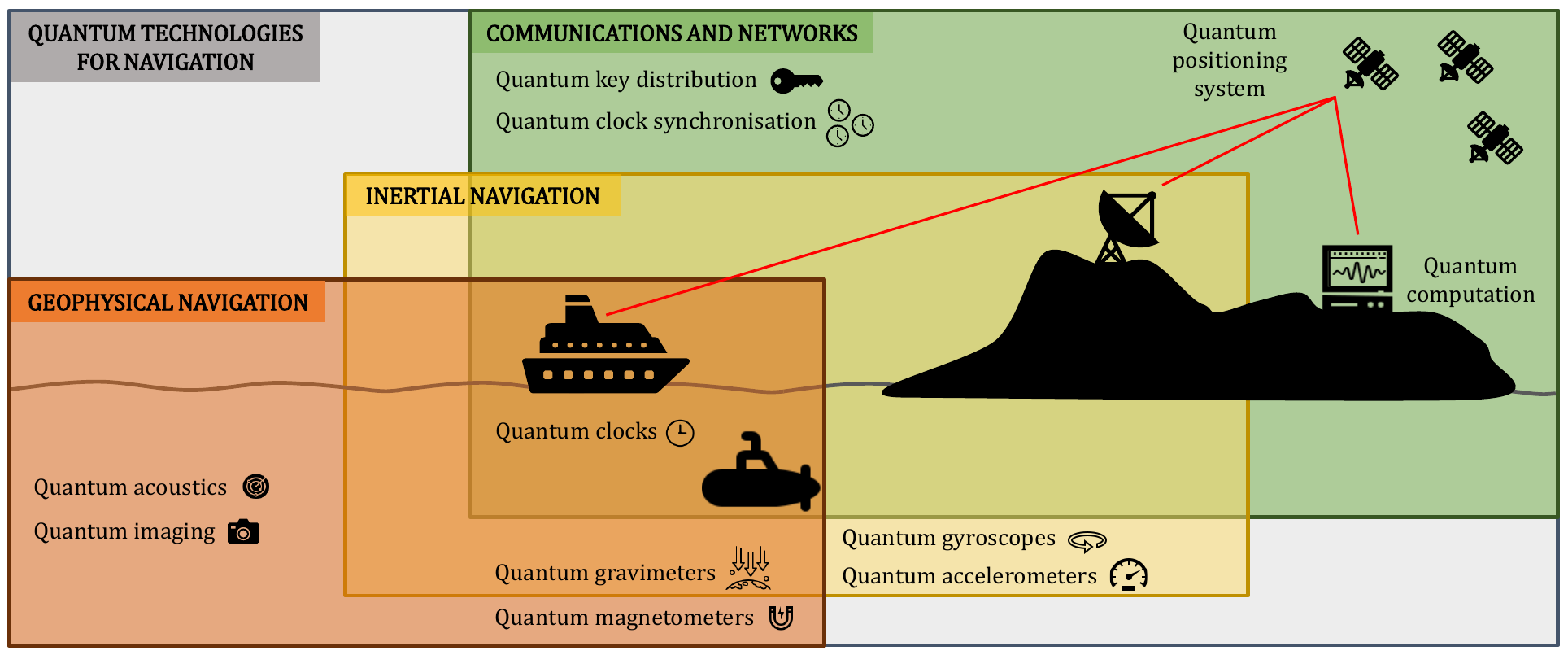}
    \caption{Outline of quantum technologies classifications. These methods could be combined together.}
    \label{fig:outline-quantum-technologies}
\end{figure*}

Quantum technologies for navigation can be categorised according to Figure~\ref{fig:outline-quantum-technologies}.

Three main categories have been identified:
\begin{itemize}
    \item Geophysical navigation: include quantum technologies that direct measure geophysical phenomena or that can be used to support geophysical navigation, e.g., terrain-based. They include quantum acoustics and imaging, quantum gravimeters and magnetometers. This category also includes quantum clocks because specific types of geophysical navigation rely on having accurate timing.
    \item Inertial Navigation: technologies that include quantum motion sensors (accelerometers or gravimeters) and rotation sensors (gyroscopes) to continuously calculate by dead reckoning the position, the orientation, and the velocity of a vehicle without the need for external references. Since precise timing is important for dead reckoning, quantum clocks also belong to this category.
    \item Communications and Networks: include technologies that use communications to provide navigational services. Some technologies, such as quantum key distribution or quantum computing, indirectly improve navigation providing security, or clock synchronisation or other services. Other technologies, such as Quantum Positioning Systems directly provide position information.
    
\end{itemize}

How quantum technology can affect navigation, and the type of navigation system used, is highly dependent on the type of operation or mission. 
In many cases, systems belonging to different categories could be combined to yield increased performance.

The rest of the paper is structured as follows:
Section~\ref{sec:state-of-the-art-maritime-navigation} discusses the current state of the art in maritime navigation.
Section~\ref{sec:a-primer-to-quantum-physical-functioning} provides an introduction to quantum physical functioning, and discusses the phenomena used in the reported technology. 
The interested readers can find additional information on the underlying quantum phenomena in Appendix ~\ref{sec:appendix} and in the references therein.
Section~\ref{sec:sensors-and-technologies-state-of-the-art} discusses the current state of the art in quantum technologies, with emphasis on maritime navigation. This section also discusses how quantum technologies might be used to overcome some of the existing navigational constraints.
Section~\ref{sec:discussions} reports key existing challenging in autonomous maritime navigation and discusses how quantum technology can be used to solve them. The section reports the key advancements required towards higher~\gls{TRL} and discusses potential timelines.
Section~\ref{sec:conclusions} draws conclusions.

\section{State of the art in maritime navigation}
\label{sec:state-of-the-art-maritime-navigation}
\begin{figure*}
    \centering
    \includegraphics[width=\textwidth]{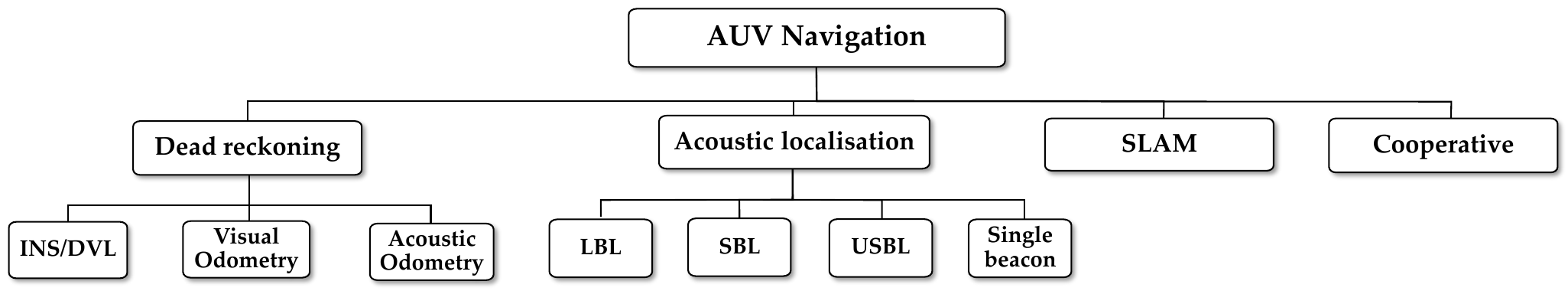}
    \caption{Classification of underwater navigation systems and sensors. Methods are often combined to provide increased performance. }
    \label{fig:navigation-categories}
\end{figure*}

Robotic marine navigation and localisation is still a challenging problem, particularly for underwater vehicles~\cite{Paull:2014}.
This is due primarily to 
the high attenuation of electromagnetic waves in seawater that makes the usage of \gls{GNSS} (a consolidated approach for terrestrial/aerial outdoor application) unusable even few centimetres below the sea surface.
Figure~\ref{fig:navigation-categories} 
reports a possible classification of underwater navigation sensors and technologies.
The rest of this section discusses the state of the art in maritime navigation and its main challenges and limitations.

\subsection{Dead Reckoning and Inertial Navigation Systems}
Traditionally, the navigation problem for \gls{UUV} is based on dead reckoning, with the vehicle trajectory estimated through the integration of the vehicle acceleration and/or velocity compensated by its orientation. 
A standard \gls{INS} solution is obtained fusing signals of inertial sensors (gyroscopes and accelerometers) with that of a compass.
Different implementations mostly differ for the specific sensors employed for measuring these quantities, and the achievable navigation performance strongly depends on the sensors used.
\textit{Gyroscopes} measure angular rates, and for marine applications ring laser or fiber optic gyroscopes are typically used. 
Angular rates are estimated based on the phase change of light as it is passed through a series of mirrors (ring laser) or through fiber optical cables.
\gls{MEMS} are low-cost inertial sensors composed of a suspended mass between a pair of capacitive plates. 
When the sensor is tilted or rotates, a difference in electrical potential is created by this suspended mass.
This can be measured as a change in capacitance and makes it possible to calculate the angular rate of the sensor.
Drifts of a gyroscope might range from \SI{0.0001}{\deg\per\hour} to \SI{60}{\deg\per\hour} for low cost \gls{MEMS} devices.
\textit{Accelerometers} measure the force required to accelerate a proof mass. Their design is typically based on pendulum, \gls{MEMS} and vibrating beams. 
Their performance range from a bias of \SI{0.01}{\milli g} for \gls{MEMS} to \SI{0.001}{\milli g} for pendulum based accelerometers.
A \textit{magnetic compass} provides a globally bounded heading reference, measuring the Earth's magnetic field vector. 
Magnetic compasses are subject to non observable bias in presence of objects that have a strong magnetic signature.
A \textit{gyrocompass} is a form of gyroscope, and uses a fast spinning wheel and the Earth’s rotation to find the true north. 
As such they are not affected by metallic objects. 
A mid-range device typically achieves accuracy performance ranging from \SIrange{1}{2}{\deg}, and their cost falls within the range of a few hundred USD.

\glspl{INS} are typically classified in four different classes, each one corresponding to a sensor grade~\cite{vectornav}. 
A change in sensor grade corresponds in a difference in performance of one order of magnitude, and for this reason, the higher the grade, the higher the cost of the sensor. 
Table~\ref{tab:vectornav-grade} summarises \gls{INS} performance based on their grade.
The navigation error, which directly depends on the integral over time of the sensor noise, is summarised in Table~\ref{tab:vectornav-error}.

It should be noted that when using an \gls{INS} in isolation and allowing its noise to double integrate, the navigation error can vary greatly depending on the sensor's quality. For a reference time span of one hour, low-cost consumer-grade \glspl{INS} can yield an error as high as \SI{39000}{\km}, while the most expensive navigation-grade sensors can achieve an accuracy of \SI{10}{\km}.

\begin{center}
\begin{table*}[ht]
\caption{\gls{INS} sensors grade classification according to Vectornav (see \cite{vectornav}, Table 3.1).}
\centering
\begin{tabular}{||p{2.2cm} p{2.5cm} p{2.7cm} p{3cm} p{2.7cm}||} 
 \hline
 Grade & 
 Accelerometer Bias (mg) & 
 Velocity Random Walk (m/s/$\sqrt{hr}$) & 
 Gyro Bias (deg/hr) & 
 Angle Random Walk (deg/$\sqrt{hr}$)\\ [0.5ex] 
 \hline\hline
 Consumer & 10 & 1 & 100 & 2 \\ 
 \hline
 Industrial & 1 & 0.1 & 10 & 0.1 \\
 \hline
 Tactical & 0.1 & 0.03 & 1 & 0.05 \\
 \hline
 Navigation & 0.01 & 0.01 & 0.01 & 0.01 \\ [1ex] 
 \hline
\end{tabular}
\label{tab:vectornav-grade}
\end{table*}
\end{center}

\begin{center}
\begin{table*}[ht]
\caption{\gls{INS} error over time by sensor grade according to Vectornav (see \cite{vectornav}, Table 3.2).}
\centering
\begin{tabular}{||p{3cm} p{2cm} p{2cm} p{2cm} p{2cm} p{2cm}||} 
 \hline
 Grade/Time & 1 s & 10 s & 60 s & 10 min & 1 hr\\ [0.5ex] 
 \hline\hline
 Consumer & 6 cm & 6.5 m & 400 m & 200 km & 39000 km \\ 
 \hline
 Industrial & 6 mm & 0.7 m & 40 m & 20 km & 3900 km \\
 \hline
 Tactical & 1 mm & 8 cm & 5 m & 2 km & 400 km\\
 \hline
 Navigation & $<$1 mm & 1 mm & 50 cm & 100 m & 10 km\\ [1ex] 
 \hline
\end{tabular}
\label{tab:vectornav-error}
\end{table*}
\end{center}

Regardless of the sensor grade, it is important to highlight that even the highest performance \gls{INS} is still of limited use for medium to long range missions, if not coupled with additional sensors that could substantially limit the error growth over time.

For better performance instead of obtaining the vehicle velocity through accelerometers integration, a direct measurement of the vehicle velocity can be obtained through a dedicated sensor as a \gls{DVL}.
The \gls{DVL} is an acoustic sensor that estimates vehicle velocity relative to the sea bottom. 
This is done by sending an acoustic pulse along a minimum of three beams, each pointing in a different direction, and measuring the Doppler shifted returns from these pulses off the sea bed to obtain velocity estimates in the device coordinate frame of reference (surge, sway and heave).
Typically, \gls{DVL} performance is quantified by a standard deviation in the range of \SIrange{0.3}{0.8}{\cm\per\s}. However, it's worth noting that utilising a \gls{DVL} requires more power and close proximity to the seabed to maintain consistent locking. 
As a result, for operations that require vehicles to move through the water column or remain far from the bottom, \gls{DVL} may not be very useful.
There are alternative methods to obtain vehicle velocity observations that involve analysing the property of some sensor signals. These methods can be based on either visual or acoustic odometry algorithms, depending on the nature of the devices being used.

Irrespective of the sensor quality and fusion algorithms used, including basic integration or more complex Bayesian approaches like \gls{EKF}, \gls{UKF}, or \gls{PF}, dead reckoning solutions are naturally associated with estimation errors that drift over time. Performance associated with these errors is typically expressed as a percentage of the distance traveled, with lower-cost sensors resulting in position errors of over 1\%, while higher-end \gls{COTS} systems integrated with \gls{DVL} can achieve position errors as low as 0.01\%, when bottom-lock is possible. 
When \gls{DVL} is unavailable, an alternative method is to infer the vehicle velocity from the propeller rotational speed, which results in much quicker error accumulation that can exceed 20\% of the distance traveled \cite{Munafo:2017jfr}.

This drift can be periodically reset through (direct or indirect) position observations.
\gls{GNSS} can be used for surface vehicles and to initialise the position of \glspl{UUV} before a dive. 
In some scenarios, \glspl{AUV} can be tasked to periodically surface to connect to \gls{GNSS} and reset their navigational errors.
It should be noted that this approach, which is frequently used in practice, results in task interruptions and puts the vehicle at increased risk. 
In addition, when vehicles are working at high depths, the cost of repeatedly surfacing can be substantial, both in terms of time and energy. 
The benefits of surfacing may be limited in such circumstances, since drift errors are unconstrained while the vehicle is transiting across the mid-water column, and currents cannot be measured properly.
For \gls{GNSS}, the position is estimated using the time-of-flight of signals from synchronised satellites, and many factors influence the accuracy of a \gls{GPS} measurements.
These include the type of \gls{GNSS} technique used, the atmospheric conditions and number of satellites in view.
The precision of the position calculated from \gls{GNSS} ranges from \SI{10}{\meter} of \gls{COTS} \gls{GPS}, to $0.3\; \text{m}-2\; \text{m}$ 
for \gls{WASDGPS}, to $0.05\; \text{m}-0.5\; \text{m}$ 
for \gls{RTK}, and to $0.02\; \text{m}-0.25\; \text{m}$ 
for post processed \gls{GNSS}~\cite{Paull:2014}.

Acoustic-based solutions, reported in the next section, are a common alternative to other positioning methods. 

\subsection{Acoustic-based Solutions}

There are several options available for acoustic localisation, including using a single beacon or multiple beacons. 
In the case of multiple beacons, the solutions are categorised according to the baseline of the sensing elements, such as \gls{USBL}, \gls{SBL}, and \gls{LBL}.
Another potential solution is to use physical cues as landmarks to implement \gls{SLAM} solutions. 
These cues can take various forms, including acoustic ones.\\
\textit{Sonars} are devices to remotely detect and localise objects in water using sound.
They can be passive, in which case they listen for sounds emitted by objects in water, or active, in which case they produce sound at specific frequencies. 
In this case, they listen for the echoes of these emitted sounds returned from remote objects in water.
Active sonars can be further divided into imaging sonars which produce an image of the seabed or ranging sonars that produce bathymetric maps.
Along-track resolution for an imaging side-scan sonar is a function of range, frequency, and water conditions. 
Cross-track performance is independent of range. 
For example, a \SI{455}{\kilo\hertz} sonar can achieve an along track resolution of \SI{10}{\centi\meter} at \SI{38}{\meter} range, and \SI{61}{\centi\meter} at the maximum \SI{250}{\meter} range. 
A sidescan sonar operating at \SI{600}{\kilo\hertz} can achieve along track resolution of \SI{5}{\centi\meter} at \SI{10}{\meter} and \SI{20}{\centi\meter} at \SI{100}{\meter}. 
In both cases nominal cross-track resolution is \SI{3.75}{\centi\meter}.
Resolution for a bathymetric sonar is on the order of \SIrange{0.4}{2}{\deg} along track and \SIrange{5}{10}{\centi\meter} cross track.
The usage of sonars for robotic navigation makes it possible to use external references to navigate.
Among others, ~\cite{Salavasidis} reports the usage of a long-range ($>$\SI{1000}{\meter}) echo sounder sonar to extend the navigation abilities of \gls{AUV} for long missions.
The system measures the sea bed altitude and uses a particle filter to find the robot most likely position, given available maps. 
Results show dramatic improvements when compared with inertial only navigation, but their quality is strongly dependent on the availability of good bathymetric maps, which are not often available.

The increasingly frequent scenario of multiple robots collaboration towards a common mission goal led to the development of navigation strategies capable of exploiting the observations of relative geometry among the cooperating agents, often made available as additional integrated services by the acoustic modems used for the necessary communication. 
This pieces of information are typically fused within the onboard navigation modules of the single agents resulting in cooperative navigation algorithms. 
It is worth noting that acoustics suffer from a number of shortcomings, ranging from limited bandwidth, low data rate and high latency~\cite{6590036}.
Environment variations strongly affect acoustic propagation, with change in water temperature and salinity that can result in dramatic variation in sound speed, multipath transmissions and varying sound channels.

The interested reader can refer to~\cite{Paull:2014} for a recent review of localisation and navigation algorithms for marine autonomous vehicles. 
Since the overall navigation performance depends not only on the available sensors but also on the specific mission requirements, the rest of this section will focus on exemplar use cases to give a more tangible representation of what can be achieved today.

\subsection{Exemplar Scenarios for Underwater Navigation}

As a first exemplar scenario, the case of polar navigation is discussed.
Low satellite coverage due to the poor visibility of geostationary (GEO) satellites has recently received a lot of attention, especially as a critical obstacle in Arctic aviation. 
The lack of robust satellite coverage coupled with a lacking quality of nautical charts (e.g., soundings are often scarce and inaccurate, resulting in the presence of unknown shoals) makes above water arctic navigation challenging, and underwater navigation even more so~\cite{SALAVASIDIS2018287, Jaakkola:2020}.
As operations at high latitudes become increasingly important, navigation and surveying in these areas has become a critical concern, particularly for underwater operations where \gls{GNSS} signals are not available~\cite{Paturel}, making inertial systems the primary method for navigation.
While inertial navigation systems are necessary for navigation at high latitudes, they also present a challenge in these conditions. 
Aligning these systems with respect to true North becomes increasingly difficult as one approaches the poles, as the physical effects on which they rely become weaker at higher latitudes.
Inertial navigation systems use the earth's rotation to align themselves with respect to true North. 
They use gyrocompasses which utilises the rotation of the earth to determine the North direction. 
As the latitude increases, this component becomes weaker and weaker, until it reaches zero at the poles, as the horizontal plane is perpendicular to the Earth's rotation rate. 
Consequently, measurement errors in the gyroscope have a greater relative impact on the heading measurement, making calibration even more crucial.
According to~\cite{Paturel}, calibration has different impacts on gyro measurements at different latitudes. Specifically, at the equator, an East bias of 0.05 degrees per hour causes a heading error of 0.19 degrees, while at 45 degrees of latitude, the same bias causes a heading error of 0.27 degrees. 
At even higher latitudes, such as 80 degrees, the same bias results in a much larger heading error of 1.1 degrees. 
This suggests that to achieve the same level of heading accuracy, a higher level of gyro performance in terms of bias repeatability is necessary at higher latitudes.

The vehicle navigation sensors must be mounted on the vehicle, and this has implications in terms of power consumption, size and cost~\cite{soton416560}.
For underwater vehicles, the inability to operate a combustion engine, due to lack of available oxygen, severely limits their power and endurance.
\glspl{AUV} must rely on batteries to power their on-board systems to provide propulsion.
New long-range vehicles are emerging with foreseen capability of 2- to 3-month-long deployments covering a range of up to \SI{1800}{\kilo\meter}~\cite{9400372}. 
While the actual performance of these \glspl{AUV} are mostly still unproven, their long-range ambitions are obtained optimising their speed and power consumption, which means that any new sensor must also be power optimised.


As another exemplar scenario, \cite{https://doi.org/10.1002/rob.21994} reports navigational performance obtained deploying long-range \glspl{AUV} travelling mid-water column in open ocean and without external support and speed over the ground measurements. 
The paper reports errors of the order of few hundred meters for a mission length of \SI{195}{\km} using a computationally optimised terrain navigation filter that used a detailed knowledge of the deployment area, i.e., a \SI{50}{\m} resolution bathymetric map.
While the algorithm is robust against map uncertainty, higher uncertainty implies higher navigation errors.

An \gls{ASW} navigation scenario is reported in~\cite{Munafo:2017jfr}.
The paper describes the deployment of an autonomous active sonar network composed of one moored gateway buoy, two wave gliders, two CMRE Ocean Explorer (OEX) \glspl{AUV} and one additional node deployed from the support ship.
The two vehicles were used as receiver of an active sonar, including on-board processing for detection, localisation, classification and tracking.
The authors report a position drift for the \glspl{AUV} of approximately 0.05\% (i.e., 0.5 m every 1 km) of the distance travelled when using a typical suite of navigation sensors including pressure sensor, \gls{DVL}, and a gyro for attitude. 
To extend the vehicle operational areas to waters deeper than $>300m$ (i.e., no \gls{DVL} bottom lock), the paper describes the usage of an acoustic network navigation system that uses two-way message exchanges between the network node to provide localisation data.
The usage of the network limits the maximum error. This depends on the turn around time of the localisation messages, which depends on the network \gls{MAC}~\cite{8084794}. Navigation errors for these network-based localisation systems range from an average of \SI{80}{\m}~\cite{Munafo:2017jfr} to less than \SI{20}{\m}~\cite{Fenucci:2022}.

Higher navigation performance are required for \gls{MCM}, exploration of archaeological sites, or more broadly for operations where the localisation of objects of interest is paramount.
Among others, a set of algorithms for the creation of underwater mosaics that can be used as visual maps for underwater navigation is reported in~\cite{GRACIAS200066}. 
Trajectory position errors of less than \SI{40}{\cm}, and maximum angular errors of \SI{3}{\deg} are reported. The system relies on carefully calibrated cameras, and on a planar map of the environment, and hence is suitable for underwater vehicles moving in the vicinity of a nearly flat ocean floor.


\color{black}{
Table~\ref{tab:quantum_vs_conventional_quantitative} presents a qualitative comparison of the anticipated performance capabilities of quantum and conventional maritime navigation technologies. 
It is crucial to recognise that assigning specific quantitative values to quantum technologies is difficult due to their early stage of development and the fact that performance metrics are influenced by various parameters, including the specific implementation and experimental conditions. 
 
The remainder of the paper examines relevant quantum-based technologies in relation to maritime navigation, offering insights into the anticipated performance enhancements and their technology readiness levels.
}
\color{black}{

\begin{table*}[h!]
\centering
\begin{tabular}{||l|l|l||}
\hline
\textbf{Technology} & \textbf{Quantum Performance} & \textbf{Conventional Performance} \\ \hline\hline
\multirow{2}{*}{Acoustics}
& High resolution, low noise; & \multirow{2}{*}{Resolution: 10 cm} \\
& Resolution: 1 mm &  \\ \hline

\multirow{2}{*}{Imaging} & Sub-wavelength resolution; ultra low SNR & \multirow{2}{*}{Resolution: $\approx 1 cm @ 10m$} \\
                                 & Resolution: $< 1mm @ 10m$ &  \\ \hline

\multirow{2}{*}{Gravimeters} & High sensitivity, low drift; & \multirow{2}{*}{Sensitivity: 1 cm/s², Drift: 0.01 $\mu$m/s²/hr}\\ 
                                    & Sensitivity: 50 nm/s²; Drift: 1 nm/s²/hr & \\ \hline
\multirow{2}{*}{Magnetometers} & Ultra-high sensitivity $<10^{-15}T$ & Sensitivity: $\approx10^{-12}T$; \\
                                    & Very low noise $< fT/\sqrt{Hz}$ & Noise $\approx pT/\sqrt{Hz}$  \\ \hline

\multirow{2}{*}{Gyroscopes} & High stability, low drift; & Accuracy: up to $0.01$°/hr;\\
                                    & Accuracy: $< 1e^{-6}$°/hr, Drift: $< 1e^{-6}$°/hr & Drift: up to 0.001°/hr  \\ \hline
\multirow{2}{*}{Accelerometers} & High precision, low noise; & \multirow{2}{*}{Precision: 10 nm/s², Noise: 100 nm/s²} \\
                                    & Precision: 1 nm/s², Noise: 10 nm/s² &  \\ \hline
\multirow{2}{*}{Quantum Key Distribution} & \multirow{2}{*}{Unconditionally secure} & Susceptible to eavesdropping \\                                     & & and cybersecurity attacks \\ \hline

\multirow{2}{*}{Clock Synchronisation} & Frequency uncertainty: 
$<10^{-18}$; & Frequency uncertainty: 
$\approx10^{-18}$; \\ 
                                    & Short term instability $< 10^{-17}$ & Short term instability $\approx 1 \times 10^{-16}$ \\ \hline

\multirow{2}{*}{Positioning Systems} & \multirow{2}{*}{Accuracy: 1 cm, global coverage} & \multirow{2}{*}{GPS: 1-5 m accuracy} \\ 
                                    & & \\ \hline
\multirow{2}{*}{Computation} & \multirow{2}{*}{Accelerated: $>$1000x speedup} & \multirow{2}{*}{Baseline (1x speed)} \\                                     & & \\ \hline
\end{tabular}
\caption{Comparison of anticipated quantum and conventional maritime navigation technology performance capabilities. Values are approximate and the actual performance of the technologies may vary based on several factors, including specific implementations and environmental conditions.}
\label{tab:quantum_vs_conventional_quantitative}
\end{table*}

\section{A primer to quantum physical functioning} 
\label{sec:a-primer-to-quantum-physical-functioning}

This section provides an introduction to quantum physics, and more specifically to those concepts that play an important role in the navigational technology discussed in the rest of the document. 
This section starts discussing the definition of state for a quantum system and of its measurements, and then it describes some of the fundamental concepts of quantum mechanics: quantisation of properties such as energy and angular momentum, superposition of multiple systems, wave-particle duality, and entanglement. 
The concepts discussed in this part are key to understand how the systems described in the rest of the paper work. 
A more in depth description of the functioning of these systems is presented in the appendix. Appendix~\ref{sec:quantum-sensing} describes quantum sensing in general, which includes all the technologies that use atomic quantum properties, for example atomic sensors and \gls{SQUID}. Appendix~\ref{sec:quantum-entangled-states}, goes into additional details on quantum entanglement describing the main methodologies for preparing entangled states: \gls{SPDC}, Ion Trap, \gls{c-QED}.

\subsection{State of- and measurement on- a quantum system}
The state of a quantum system is a mathematical entity, which we will indicate with $|\Psi\rangle$ (using the bracket notation).
It collects all the possible outcomes of a measurement on it and associates a probability density to each of them~\cite{griffiths2018}. 
One fascinating and distinctive feature of quantum mechanics is the notion of superposition, which means that a quantum system can exist in a combination of all the possible states described by its probability distribution simultaneously, until it is observed or measured.
The material properties of objects do not have a definite value or existence until they are observed or measured. 
The act of measurement causes the system to "collapse" into one of its possible states, with the outcome determined by the probabilities associated with each state.
The superposition principle is a fundamental concept in quantum mechanics that states that if a system has $n$ possible states represented by $\psi_1,\psi_2,...,\psi_n$, then any linear combination of these states, $\psi = c_1\psi_1+c_2\psi_2+...+c_n\psi_n$, must also represent a valid and possible state of the system. 
The coefficients $c_1, c_2,...,c_n$ are complex numbers that determine the relative weights of each state in the superposition and have a probabilistic interpretation. 
When the state is measured, the probability of finding the system in one of its possible states is proportional to the magnitude squared of the corresponding coefficient.
Upon measurement of a quantum system in a superposition of states $\psi = c_1\psi_1+c_2\psi_2+...+c_n\psi_n$, the probability of obtaining a particular state $\psi_i$ is given by the square of the absolute value of its coefficient $c_i$, denoted as $|c_i|^2$. 
The probabilities of all possible outcomes must sum to one, which implies that $\sum_{i=1}^{n} |c_i|^2=1$.

One simple example of a quantum system is a qubit, which is the quantum equivalent of a classical bit (denoted as 0 or 1). In the case of a qubit, there are only two possible states, represented as $|0\rangle$ and $|1\rangle$, and the coefficients of the superposition are $c_1$ and $c_2$ respectively, with $|c_1|^2+|c_2|^2=1$. The probabilities of measuring the qubit in the states $|0\rangle$ and $|1\rangle$ are $|c_1|^2$ and $|c_2|^2$ respectively.
It is then possible to describe describe its state before measurement as a linear combination of the two possible states $|0\rangle$ and $|1\rangle$, such that

\begin{equation}
|\Psi\rangle=c_1|0\rangle+c_2|1\rangle
\end{equation}

This state is a superposition of the two basis states and can represent any state of the qubit.

Upon measurement of the qubit, it collapses into one of the two basis states $|0\rangle$ or $|1\rangle$. 
The probability of the qubit collapsing into state $|0\rangle$ is $|c_1|^2$, while the probability of collapsing into state $|1\rangle$ is $|c_2|^2$. Since the sum of these probabilities must equal 1, the qubit must collapse into one of the two states.

\subsection{Quantisation of Energy}
Unlike the classical physics, quantum systems can possess only certain discrete energy values and not a continuum of values~\cite{griffiths2018}. 
One example of quantised energy is the energy levels of an electron in an atom. 
The electron can only occupy certain discrete energy levels, and transitions between these levels result in the emission or absorption of photons with specific energies.
Quantisation is a fundamental concept in quantum mechanics that is not limited to energy. 
Other physical quantities such as angular momentum, electric charge, and magnetic moment can also be quantised in quantum systems.


\subsection{Wave particle duality}
Wave-particle duality~\cite{Cohen-Tannoudji} is another fundamental concept in quantum mechanics that describes the behaviour of particles at the quantum level. According to this concept, particles such as electrons and photons can exhibit both wave-like and particle-like properties, depending on how they are observed or measured.

In other words, as proposed in 1924 by the french physicist Louis de Broglie, each individual massive particle has a wavelength $\lambda$ directly proportional to the Plank constant $h$ and inversely proportional to its momentum $p$
$$ \lambda = \frac{h}{p} = \frac{h}{mv}$$
in which $m$ is the particle mass and $v$ its velocity.

The same characteristics have been observed not only in elementary particles, but also in compound particles like atoms and even molecules, generating the so-called matter wave.

\subsection{Qubits}
A qubit, or quantum bit, is the quantum version of the classical binary bit, and is considered the fundamental unit of quantum information~\cite{nielsen_chuang_2010}. 

Qubits are typically associated with microscopic systems such as the two opposite polarisations of a photon, the two alignments of a nuclear spin in a uniform magnetic field, or the two energy levels of an electron orbiting a single atom.
 

In the case of nuclear spins, the two states are generally referred to as \textit{spin up} $|0\rangle=\ \uparrow$ and \textit{spin down} $|1\rangle=\ \downarrow$. 
For photon polarisation, we have horizontal $|0\rangle=\ \leftrightarrow$ or vertical $|1\rangle=\ \updownarrow$. 
However, in some circumstances it may be useful to use a different basis to describe the system and take advantage of the superposition principle, using a specific linear combination of $|0\rangle$ and $|1\rangle$. 

Techniques, such as \gls{QKD} (described in ~\ref{sec:qkd}) and quantum teleportation, use the so-called Hadamard basis: 

\begin{align}
    |+\rangle=\sqrt{\frac{1}{2}}(|0\rangle+|1\rangle)=\ \neswarrow\label{eq:hadamard+}\\
    |-\rangle=\sqrt{\frac{1}{2}}(|0\rangle-|1\rangle)=\ \nwsearrow\label{eq:hadamard-}
\end{align}

which corresponds to a $45^{\circ}-135^{\circ}$ polarisation.\\
Typically, a two-level quantum system can be visualised in the so-called Bloch sphere~\cite{Zienau_1975}, shown in Figure~\ref{fig:bloch}. The two states are the vector pointing along $+z$ and $-z$ and all the vectors pointing in the other directions are the superposition states. In presence of an external field (e.g., magnetic field acting on a spin), a system prepared for example in the $|0\rangle$ state, evolves for a certain time (it performs a rotation $\omega$ in the Bloch sphere) until it reaches its final state. 
This will be a generic combination of $|0\rangle$ and $|1\rangle$ and will be represented by a vector in the Bloch sphere with components proportional to the associated probability densities.

\begin{figure}[htp]
\centering
\includegraphics[scale=0.5]{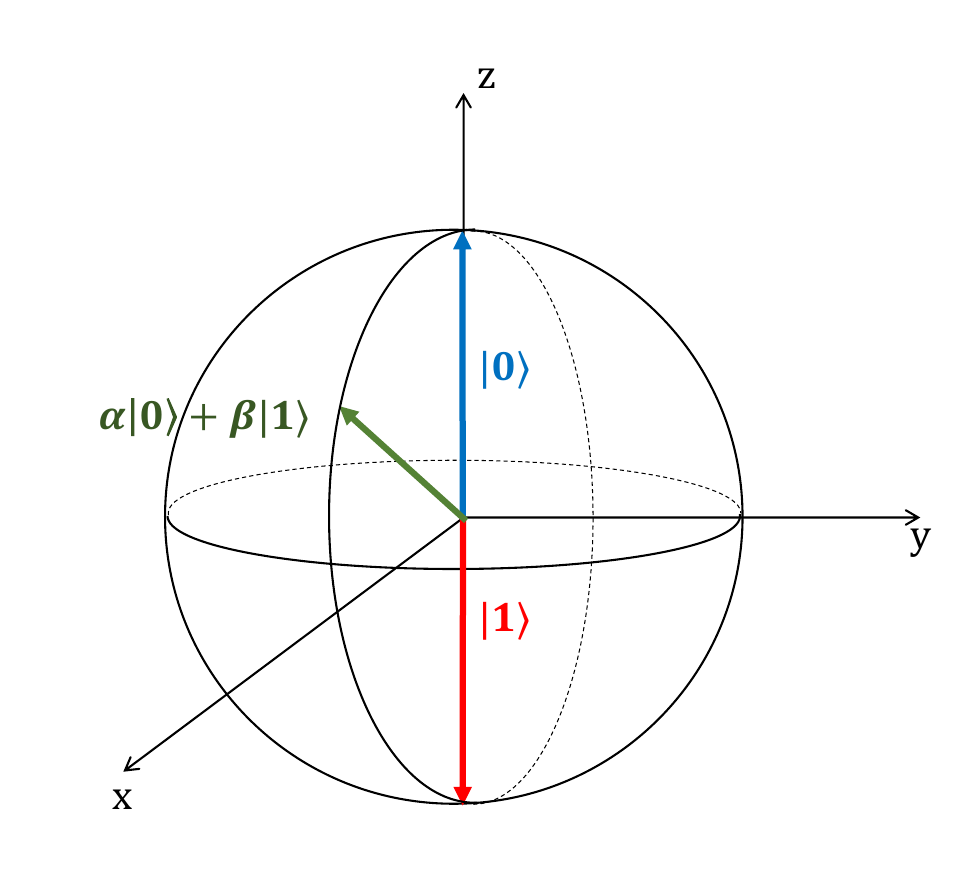}
\caption{Qubit representation in the Bloch sphere. The two basis elements ($|0\rangle$ and $|1\rangle$) are the vector pointing along $+z$ and $-z$. All the other vectors represent the linear combinations of the basis vectors, i.e., the superposition states.}
\label{fig:bloch}
\end{figure}

The Appendix~\ref{sec:atomic-sensors} describes a process which uses the first two energy levels of an atom. These levels serve as a qubit, and the process forms the basis of atomic interferometry.

\subsection{Quantum entanglement}
Entanglement is a unique property of quantum mechanics that has no classical analogue. 
It refers to a phenomenon where the quantum state of two or more particles cannot be described independently, even when they are physically separated. 
This means that the state of one particle is intrinsically linked to the state of the other particle, and any measurement made on one particle instantaneously affects the state of the other, regardless of the distance between them.

Given that each qubit can be in one of two possible states, $|0\rangle$ or $|1\rangle$, and the overall state of the pair is a combination of the individual states, the state of a pair of qubits can be described by one of the four boolean states $|00\rangle$, $|01\rangle$, $|10\rangle$, or $|11\rangle$ (or any combination of these).
If the two qubits are separable (i.e., their state is not entangled), their state can be expressed as the tensor product of the states of the single photons, for example:

\begin{equation}
    \sqrt{\frac{1}{2}}\left(|00\rangle+|01\rangle\right) = |0\rangle \otimes \sqrt{\frac{1}{2}}\left(|0\rangle+|1\rangle\right) = |0\rangle\otimes|+\rangle=\ \leftrightarrow\neswarrow.
\end{equation}
where $|+\rangle$ is defined in (\ref{eq:hadamard+}).

If however their state cannot be expressed as the tensor product of the states of the single photons, as in the case of:
\begin{equation}
    \sqrt{\frac{1}{2}}\left(|00\rangle+|11\rangle\right) 
\end{equation}

they interact in such a way that the quantum state of each subsystem cannot be described without including information about the others, and their are said \textit{entangled}. 

This correlation is non-local, which means that it cannot be explained by any local interaction between the particles, and continues to be valid even when the particles are spatially separated from each other.
When two particles become entangled, the state of one particle cannot be described independently of the state of the other particle, even if they are far apart.

The non-local character of the entanglement is an interesting aspect in the context of quantum technology. 
It enables the design of instruments with an arbitrary spatial distribution, where some components can operate and make measurements on one subsystem, leading to a remote influence on another distant subsystem.

Using these properties, entanglement can be used for a range of applications including quantum cryptography, quantum computation (see \ref{sec:quantum-computing}), quantum communication (\ref{sec:underwater-quantum-communications}), quantum imaging (\ref{sec:quantum-imaging}), etc.
 
\section{Sensors and technologies: state of the art} 
\label{sec:sensors-and-technologies-state-of-the-art}

This section discusses the current state of the art of a range of quantum sensors and technologies that can be used for maritime navigation.
Results are reported on what has been already achieved through experiments, prototypes or commercially available technologies.
Section~\ref{sec:discussions} will later discuss what might be possible in the future through a further evolution of the technologies.

\subsection{Quantum Computing}
\label{sec:quantum-computing}

Quantum computers perform calculations using qubits, which, as described in the previous section, are typically realised through subatomic particles (e.g., the spin up/down of an electron or the horizontal/vertical polarisation of a photon).
The concept of superposition of multiple states, together with that of coherence and entanglement, are the foundation on which quantum computing is based
to obtain an exponential increase in their computing power over classical computations.

The unmeasured quantum state of a qubit simultaneously exists in all the states specified by its underlying probability distribution. Additionally, changing the state of an entangled qubit will instantly affect the state of the paired qubit. These properties give quantum computers the potential to process exponentially more data than classical computers.
As discussed in~\cite{10.2307/3560059}, representing a superposition of $2^n$  levels classically requires assigning a physical property such as an electronic energy level to each individual quantum state within the superposition.
However, because the number of possible quantum states in a superposition grows exponentially with $n$, the number of physical properties that would need to be assigned would also grow exponentially. This means that the amount of physical resources (such as memory or processing power) required to represent and manipulate these quantum states using classical computing methods would also grow exponentially with $n$.

In quantum systems however, quantum entanglement makes it possible to use $n$ two-level systems to represent a general superposition of $2^n$ levels and the amount of the physical resource (that defines the levels) will grow only linearly with $n$. 
Leveraging specialised algorithms and hardware ~\cite{Arute:2019} makes it possible to sample the output probabilities and hence explore exponentially large states. 

As a result, quantum computers can now run interesting programs~\cite{AQS}.
In \cite{Arute:2019}, the authors reported on a 54-qubit processor comprised of fast, high-fidelity quantum logic gates designed to perform benchmark testing.
The quantum computer performed the target computation in 200 seconds, a value that the researchers compared to 10,000 years to produce a similar output using the world’s fastest supercomputer.
While this specific result might not have a short term practical applicability, it was the first experiment where a quantum computer surpassed state of the art classical computers.

A quantum algorithm for approximate combinatorial optimisation is reported in~\cite{https://doi.org/10.48550/arxiv.1411.4028}.
The development of quantum annealers able to tackle hard optimization problems with rugged energy landscapes is reported in~\cite{Denchev_2016}. 
The paper reports how exploiting tunneling, a quantum phenomenon that involves microscopic objects and allows them to traverse tall and narrow energy barriers (see Appendix~\ref{sec:squids}), quantum algorithms are able to run more than 108 times faster than the equivalent simulated annealing algorithms running on a single core.
Although the results of quantum computing are promising, more work is needed to make it a practical technology.

\subsubsection{Challenges and current developments}
Creating a functional quantum computer requires maintaining an object in a superposition state for an extended period to perform various operations. However, when a superposition comes into contact with materials that are part of the measurement system, it loses its in-between state, a phenomenon known as decoherence, and becomes a classical bit. To be useful in practice, quantum computers must decrease hardware errors, improve the generation and management of qubits, and reduce decoherence while still making the qubits easy to read. This could be achieved through more robust quantum processes or by developing new error-checking methods.
Theoretical predictions on quantum limits suggest that there are tighter limits to what quantum computers can achieve due to noise corrupting the computation that cannot simply be reduced~\cite{Kalai}.
Quantum systems are inherently noisy due to interactions with the environment. This noise can corrupt the quantum states being used for computation, leading to errors in the results of the computation. 
Benchmarks have been recently designed to measure capability directly, overcoming some of the limited flexibility and scaling properties of early benchmarks and providing a possible way to better predict performance of useful programs~\cite{Proctor_2021}.
This will be a key enabler for the design of next generation quantum algorithms and facilitate the embedding of problems of practical relevance.

Quantum computers might also not fully replace conventional computers, with classical ones that could retain an economical edge for tackling most problems. 
A recent assessment of the quantum computing technology place the highest TRL to 5~\cite{trl-level-doc} when using a particular type of qubit called superconducting qubits~\cite{nielsen_chuang_2010}, while different technologies (e.g., trapped ions -  Appendix~\ref{sec:ion-trap}) that have the promise to improve some of the existing limitations (e.g., longer qubit life) are further behind along the technology readiness scale.
The recent tendency is to leverage quantum computing to augment exist computational workflow with applications ranging from simulation, optimisation, and machine learning. 
This can be done, for example, using quantum computers to narrow the range of possible solutions of the problem and then applying classical computer to fully solve it.
From a navigation perspective the ability to leverage quantum computation can represent an advantage.
Early attempts include the development of quantum-inspired algorithms that leverage the mathematical concepts of qubits and superposition to solve large scale optimisation problem for robotic path planning~\cite{5715406, doi:10.1142/S0218126620501224}.
Similarly, some preliminary work~\cite{Lu2020} is applying quantum concepts for particle filtering, using qubit states to model particle evolution.
While these works do not make usage of quantum mechanics, but rather of the underlying mathematical concepts, they seem to point towards algorithms that could take advantage of quantum computations in the future.


\subsection{Quantum inertial sensors}
\label{sec:quantum-intertial-sensors}
Inertial sensors can be classified according to their applications, and typically they are divided in sensors for inertial navigation (accelerometers and gyroscopes) and sensors to measure the gravitational acceleration (gravimeters)~\cite{vectornav:resources}.
Gravimeters and accelerometers measure accelerations, but their specific objectives result in different implementation setups.
Gravimeters are high sensitivity single axis accelerometers that measure the Earth gravity acceleration $g_0$. Given their specific objective they are highly specialised sensors to target signals with intrinsic slow dynamics.
Accelerometers are the primary sensors to measuring inertial acceleration, or the change in velocity over time.
They have less stringent sensitivity requirements when compared to gravimeters but they need to measure highly dynamic signals.
Accelerometers are typically installed on movable platforms (e.g., \gls{AUV}) and for this reason consideration on power consumption, robustness to challenging environments and size, play an important design role. 



The next subsections go into the details of how quantum-technologies can be used to overcome some of the existing limitations of classic inertial sensors~\cite{templier2021three}, with atom interferometry, which uses cold atoms acting as free falling proof masses in a vacuum, and \gls{CAI} as the most promising quantum technologies in this sector. Both technologies are explained in the Appendix~\ref{sec:quantum-sensing}, together with the various interferometry concepts mentioned here, as matter-wave interferometer, Mach-Zehnder interferometeric scheme and so on.
New quantum sensors seem to be able to dramatically improve the expected residual measurement error, and this has a direct impact on the quality of the navigation solution that can be obtained from \glspl{INS}.
 \gls{INS} calculates an estimated trajectory (vehicle pose as a function of time) integrating over time inertial signals (e.g., accelerations need to integrated twice to obtain a position).
 The limiting factor in classical \glspl{INS} is due to the (double) integration of the measurement errors, which corresponds to have an inherent error drift, which effectively constraints the usage of \glspl{INS} navigation solutions to short-term operations only.
 As will be discussed in the next subsections, the availability of quantum sensors with errors orders of magnitude smaller than existing systems, would hold the promise to enable long-term inertial navigation.\\

\subsubsection{Inertial Measurements}

The first experimental evidence of inertial measurements based on \gls{CAI} is reported in~\cite{kasevich1992measurement} exploiting a Sodium atomic fountain in a Mach-Zehnder interferometric scheme.
Building on these results, in 2006, the first \gls{CAI} gyroscope was demonstrated using Caesium atoms. 
The \gls{CAI} gyroscope was reported (see~\cite{durfee2006long}) to have a bias stability 300 times better than those obtainable with state of the art commercial navigation grade gyroscopes (laser ring and fiber optics), and 1000 times better  angle random walk (the noise component perturbing the output of a gyroscope).
Also in 2006, \cite{canuel2006six} reported experimental results from a unique instrument based on Caesium atoms that was able to simultaneously measure three-axes accelerations and three-axes angular rates. 
The drawback of these systems lies in their intrinsic high sensitivity, which also means that they are sensitive to environmental variations, specifically, vibrations and temperature.
To limit their environmental exposure, existing systems are typically run under strict laboratory conditions to guarantee the necessary isolation.
The first experimental demonstration of an inertial sensor based on matter-wave interferometry mounted on an aircraft was reported in~\cite{geiger2011detecting}. 

The matter-wave interferometer using ${}^{87}$Rb atoms was installed aboard the Novespace A300 – 0g aircraft operating from Bordeaux airport, France. The plane, which is, in effect, a flying laboratory, is able to perform parabolic flights alternating ballistic (zero gravity) trajectories and standard gravity flights.
The matter-wave interferometer was able to detect the aircraft acceleration with respect to inertial frame attached to the interrogated free-falling atoms. 
To be mounted on an airplane, the system was considerable more compact than other systems available at the time.
The reduction in size was enabled by telecom-based laser sources, which also had the advantage of guaranteeing high-frequency stability.

Acceleration fluctuations along the direction of the wings demonstrated to be three orders of magnitude higher than what commonly found with laboratory matter-wave interferometry inertial sensors.
Measurements are based on the correlation between the atom interferometer and the mechanical accelerometers fixed on the system retroreflecting mirror. 
The overall instrument can thus be seen as a hybrid sensor that benefits from the complementary property of its mechanical components (able to measure large accelerations) and of its atomic elements (which provide higher resolution).

Another compact instrument for high precision sensing of rotations and accelerations was reported in~\cite{muller2009compact}. 
The system employed an interferometer working simultaneously with two atomic sources that were injected with opposite velocity directions.
In its usual configuration the system has a total length of \SI{90}{\cm}, and it is integrated on a non-magnetic optical breadboard with a dimension of \SI{120}{}$\times$\SI{90}{\cm\squared}, and that makes the instrument small enough to be moved during experimental campaigns.

In 2014, a research carried out at the Sandia National Laboratories of Albuquerque (New Mexico, USA) proposed to recapture the atoms after the interferometer sequence with the goal of reducing the dead time between measurements related to the cooling process~\cite{rakholia2014dual}.
This resulted in an increased measurement frequency for the high-data-rate atom interferometer, demonstrating simultaneous measurements of acceleration and rotation. 
Results were for dual-axis and sampled at \SI{60}{\hertz}, with an achieved sensitivity of \SI{0.9}{\micro g\per\sqrt{\hertz}} for accelerometers and \SI{1.1}{\micro\radian\per\sqrt{\hertz}} for gyroscopes running at the same rate. Based on these results, the authors suggested that their inertial sensors are ready to move beyond the walls of the protected environment of the laboratory.
A hybrid accelerometer which takes advantages of conventional and atomic sensors in terms of bandwidth (DC to \SI{430}{\hertz}) and long term stability has been proposed in~\cite{lautier2014hybridizing}.
The classical accelerometer is used for a real-time comparison of the interferometer phase shift and this makes possible to improve performance without the need of an isolation platform.
Moreover, the atomic sensor enables to servo-lock the output of the conventional accelerometer, suppressing its drifts to follow gravity changes, and permitting the use of the classical accelerometer in a fluctuating thermal environment. 
Results reported after calibration to remove the acceleration offset (bias) show a navigation error of less than \SI{1}{\m} after \SI{4}{\hour}.
It is important to highlight that this result is estimated integrating over time the residual bias and that no actual navigation was experimented (and hence the expected error is not expressed as a function of the distance travelled by the platform that may host the device).
The reader can compare this result with some of the \gls{INS} metrics reported in Table~\ref{tab:vectornav-error}, Section~\ref{sec:state-of-the-art-maritime-navigation}.

Improved sensitivity and accuracy can be achieved using beam splitters and mirrors using composite light pulses~\cite{berg2015composite}.
The resulting \gls{SCI} differs from the \gls{MZI}, and it does not require a preparation step before the interferometer sequence. 
The solution is a combination of both the Bragg- and Raman-type concepts, both dealing with light-matter interaction. Without entering into the details, Bragg's law describes how a crystal lattice scatters electromagnetic radiation (such as x-ray, gamma ray or matter waves), and provides the angles for coherent and incoherent scattering, which together describe the pattern formed by the reflected radiation. 
Raman spectroscopy, instead, deals with vibrational and rotational modes of a system and Raman scattering in particular describes inelastic scattering of light by matter. It is described in detail in Appendix~\ref{sec: Atom interferometry}.

A reported approach to target performance improvement in terms of sampling rates and sensitivity has been based on the use of interleaved operations. \cite{savoie2018interleaved} demonstrated the simultaneous interrogation of three atomic clouds within a cold atom gyroscope achieving a sampling rate of \SI{3.75}{\hertz} with an interrogation time - time spent by the atoms inside the interferometer - of \SI{801}{\ms} and ensuring, at the same time, a high inertial sensitivity. 
The paper discusses how the method could be extended to atom interferometers more broadly and how it might pave the way for high-bandwidth and high-sensitivity cold-atom inertial sensors.

Given that weaknesses of quantum accelerometers tend to align with strengths of classical ones and vice-versa, having a hybrid setup has been proposed by more than one research group. 
\cite{wang2021enhancing} proposed a maximum likelihood estimator to fuse measurements and to obtain a hybrid setup for inertial navigation. 
Results are reported in simulation only and show promising performance in terms of extension of the dynamic range of quantum accelerometers and minimisation of drift of classical accelerometres. 
Similar results are obtained in~\cite{cheiney2018navigation}. In this case, iXBlue and LP2N (iXAtom laboratory) use an \gls{EKF} to validate in post-processing a static hybrid system. Data are obtained in experimental conditions for a vertical degree of freedom with artificially injected disturbances representative of a mobile scenario (vibrations and heating).
Further work was reported in~\cite{templier2021three} where a 3-axis hybridized accelerometer was proposed as a first step towards multi-axis atomic devices for inertial navigation. 
The sensors were statically validated in a room with no control on environmental conditions.
Finally, \cite{weidner2018experimental} experimentally demonstrated a shaken lattice interferometry as an alternative to state-of-the-art atom interferometry techniques.

A significant number of techniques are under active investigation, with interesting results suggesting improvements of orders of magnitude with respect to existing classical methods.
Results have been reported only in limited laboratory conditions. Device miniaturisation and the extension of their dynamic ranges for operational use (i.e., out of laboratory conditions) represent the main engineering challenges in the years ahead to successful technology ransition~\cite{krelina2021quantum}.\\


\subsubsection{Gravity Measurements}
Instruments for gravity measurements include spring and corner cube gravimeters, and \gls{CAI} gravimeters.
The former sensors are small in size and easily portable, whereas \gls{CAI} gravimeters are still bulky and complex, and need a suitable non-disturbing environment to operate.
Significant weight, size and energy absorption reductions, together with an increased robustness against noise and external disturbances are needed for \gls{CAI} gravimeters to be used in operations, especially aboard mobile maritime platforms. 
Recent cold-atom based gravimeters are smaller and transportable (i.e., it is possible to transport and install them from one site to a different one) but still not suitable for usage on mobile platforms. 
Moreover, they still rely on environments with little or no external disturbances to achieve nominal performance.
Two relatively compact gravimeters, produced by AOSense \cite{aosense} and $\mu$quans \cite{muquans}, are commercially available.
 $\mu$quans reports a sensor head size with an height of \SI{70}{\cm}, a diameter of \SI{38}{\cm}, and a weight of \SI{25}{\kg}. 
 The laser and electronics are {\SI{100}{\cm}x\SI{50}{\cm}x\SI{70}{\cm}} and weigh \SI{75}{\kg}; the overall system absorbs under normal operation \SI{250}{\watt}. 
 Figure~\ref{fig:muquans} shows the $\mu$quans gravimeter as presented on the company's website.
 
\begin{figure}[htp]
\centering
\includegraphics[width=0.5\textwidth]{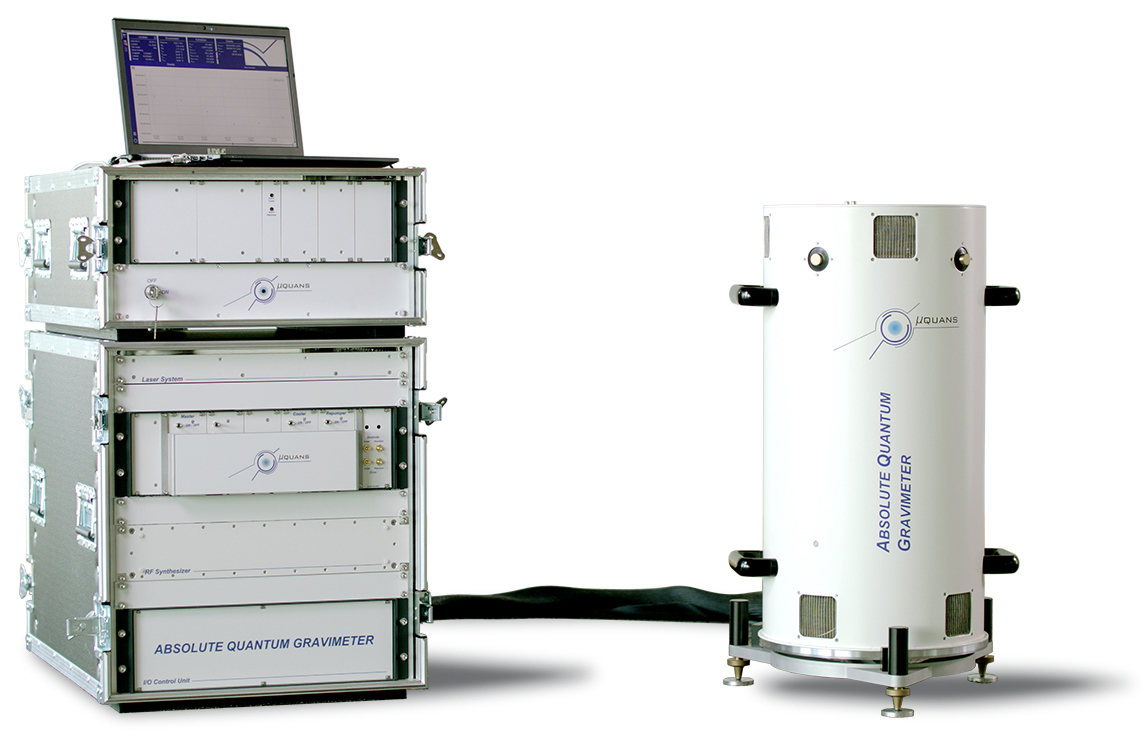}%
\caption{Commercial gravimeter produced by $\mu$quans as shown on the company's website.}
\label{fig:muquans}
\end{figure}
Examples of similar, but less compact, devices are available as research instruments \cite{travagnin2020physics}.

The availability of gravimeters able to provide extremely precise gravity measurements can be key for navigation. 
Subtle gravity variations that are geographically specific, can be detected and potentially used to plot the vessel’s position matching the measured fluctuations to a gravity map of the globe (see Figure~\ref{fig:nasa-gravity}). 
Simulations suggest that such a technique could get to an accuracy of around~\SI{10}{\meter}~\cite{nasa-gravity}.

\begin{figure}[htp]
    \centering
    \includegraphics[width=0.5\textwidth]{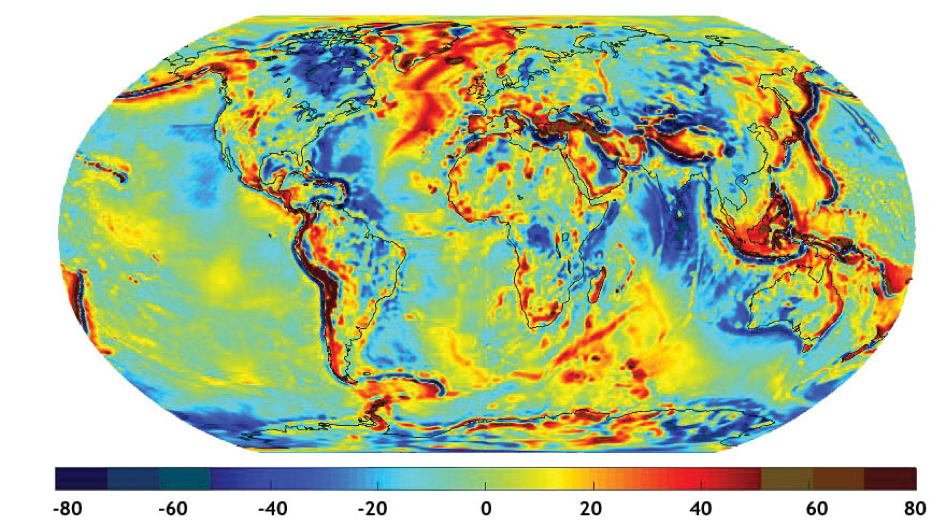}
    \caption{Gravity anomalies from ten years (2003-2013) of Gravity Recovery and Climate Experiment (GRACE) data and four years of GOCE data (GGM05G Model) (see~\cite{nasa-gravity})}
    \label{fig:nasa-gravity}
\end{figure}

\subsection{Quantum magnetometers}
\label{sec:quantum-magnetometers}
Magnetometers also play an important role in navigation and are used mainly for two reasons:
\begin{itemize}
    \item Attitude estimation: a multi-axes device is employed to measure the projection of the Earth Magnetic Field along three orthogonal directions, similarly to what a compass does. 
    A measurement of the North direction with respect to the platform which hosts the device is thus complemented with the signal of inertial sensors (accelerometers and gyroscopes) to obtain the three dimensional estimation of the attitude within an \gls{AHRS}.
    \item Geophysical navigation: measurements of local magnetic field and of its space variation are compared with known maps to correct the platform estimated position, for example maximising the likelihood of the measurements (see ~\cite{Salavasidis} for a similar approach for terrain-based navigation).
\end{itemize}

Quantum magnetomers can achieve higher levels of sensitivity than classical magnetometers. Quantum phenomena behind their functioning are reported in Appendix~\ref{sec:atomic-magnetometer}.


The current standard for commercial quantum magnetometers in laboratory applications is the \gls{SQUID}. \gls{SQUID} magnetometers are very sensitive magnetometers used to measure extremely subtle magnetic fields, based on superconducting loops containing Josephson junctions (see Appendix~\ref{sec:squids}).
\glspl{SQUID} are sensitive enough to measure fields as low as \SI{5e-14}{\tesla} \cite{nasa-squid}, with noise levels as low as \SI{3}{\femto\tesla\per\sqrt\hertz}~\cite{4277368}.
For comparison, a typical refrigerator magnet produces 0.01 Tesla ($10^{-2}\ T$).
They need cryogenic refrigeration to operate.

Recently, atomic vapour cell magnetometers have been demonstrated to perform better than \glspl{SQUID}, with the advantage of working at room temperature \cite{korth2016miniature}. 

The first experimental demonstration of phenomenons that today are used in \glspl{OAM} dates back to Michael Faraday's discovery of the magneto-optical effect of light with matter.
This provided evidence of a relationship between light and electromagnetism (Faraday effect, Faraday rotation or the \gls{MOFE}).
In Faraday's work the observation of the rotation of the polarisation of the light generated by an Argand lamp was observed when light was transmitted parallel to the magnetic field through a block of glass.
On the basis of the phenomenon of polarised light angular momentum transfer to atoms of Mercury demonstrated by Dehmelt \cite{dehmelt1956paramagnetic}, \cite{dehmelt1957modulation} proposed that magnetic field strength could be observed exploiting the Larmor precession.
In 1957, Bell and Bloom developed the first alkali-metal magnetometer integrating the setup of Dehmelt for optical pumping with an additional beam (a cross beam) to evaluate the precession of the Sodium atoms \cite{BellBloom1957}. 
\glspl{OAM} usually use vapours of alkali metal for sensing. The use of Helium has been applied for space applications~\cite{budker2007optical}. 
Regardless of specific implementation solutions, \glspl{OAM} use light to observe the magnetic moments of atoms as influenced by the presence a magnetic fields. They represent a modern day direct translation of the device used by Faraday: it consists of a first phase of spin polarisation of the atom sample by transferring angular momentum; then a torque is applied to the sample subjected to a magnetic field resulting in a precession at the Lambert frequency; this, in turn, results in a modification of the spin that is detected to derive the strength of the magnetic field itself.
Different implementation solutions for the various phases of the process have been developed to address specific requirements, e.g. sensitivity, operating temperature, source of light, etc.~\cite{hrvoic2005brief}. 
The solution that demonstrated the highest sensitivity, i.e., the ability to detect the weakest magnetic field was proposed at Princeton \cite{kominis2003subfemtotesla,allred2002high} and named \gls{SERF}.
\glspl{SERF} magnetometers are designed to work with a high atomic density in the near-zero field to overcome the limit to the sensitivity due to spin exchange collisions between the atoms. 
Under the conditions of the proposed setup, the obtained spin exchange is faster than the precession frequency providing an almost constant average spin. 
The potential sensitivity reported is less than \SI{0.01}{\femto\tesla\per\hertz}.
Recent work \cite{hussain2018application} has proposed an optically-pumped \gls{RF-OAM} based on Rubidium vapour that promises to be portable, operable at room temperature and in unshielded environments. 
Finally, recent work is showing that superconducting qubits that are used in quantum computers can be used to measure electric and magnetic fields \cite{degen2017quantum}.


As discussed above and similarly to gravimeters, the precision and accuracy promised by quantum magnetometers might make it possible to sense and map Earth’s magnetic field and its gradient to levels that are not possible today. 
The availability of high resolution map and of sensors that can precisely locate within these maps can be a critical enabler for high-accuracy navigation. 
Moreover, the reported increase in the detected magnetic precision suggest that it should be possible to estimate vehicle attitude at higher latitudes, and hence alleviate some of the existing navigational constraints.


\subsection{Underwater quantum communications}
\label{sec:underwater-quantum-communications}

Underwater quantum communications are receiving increased attention recently, mostly as a byproduct of the inherent interest in secure communications, with recent work investigating how seawater might affect quantum properties.
\cite{Shi:2015} performed Monte Carlo simulations to investigate how optical absorption and scattering properties of underwater media can affect \gls{QKD} (see Section~\ref{sec:qkd}) channel models, and communication performance measured in terms of quantum bit error rate, and sifted key generation rate. 
A set of experiments performed in the Yellow Sea and reported in ~\cite{Ling:2017}, demonstrated that polarisation quantum states including general qubits and entangled states can  survive after travelling through seawater, verifying the presence of a quantum channel over \SI{3.3}{\meter}.
A similar experiment performed with similar communication ranges is reported in~\cite{Zhao:2019}.
The authors show that the majority of the photons are unscattered, implying high polarisation properties can be maintained with high reliability.
Results also show that under those experimental conditions \gls{QKD} can be performed with a quantum bit error rate less than 3.5\%, with different attenuation coefficient.

Longer ranges of up to \SI{55}{\meter} have been reported in~\cite{Hu:2019} in a \SI{300}{\meter} x \SI{16}{\meter} long tank with a maximum depth of \SI{7.5}{\meter}. 
Photons emitted by the transmitter are steered into water by a wireless-controlled cradle head and finally guided back into free-space air before reaching the receiver. 
The link was a relatively complex channel composed of two air-water interfaces and a \SI{55}{\meter} long underwater channel at \SI{1.5}{\meter} depth.
A \SI{532}{\nano\meter} continuous laser with \SI{100}{\milli\watt} power is used at the transmitter side, and then amplitude modulated to attenuate it to the single-photon level ({and with an intensity required by the standard decoy-state protocol}).
The obtained output power can be as low as \SI{0.1377}{\nano\watt}. 
The authors highlight a number of challenges, including in-water laser alignment and the need of additional lenses to expand the light beam and increase its Rayleigh length for better stability and transmission efficiency, and the 
difficulties in retrieving photons from strong background noise in high-loss conditions.
Channel loss is obtained comparing transmitted and received powers, with a measured attenuation coefficient $\alpha=\SI{0.16}{\per\meter}$, similar to those characterising coastal seawater.
The total loss of the system is reported to be about \SI{40}{\dB}.
Based on the achieved attenuation coefficient, the authors hints to the possibility of achieving longer distances but no evidence or further details are provided.


Recent experiments carried out in Ottawa River in Canada have started to analyse the propagation of optical beams possessing different polarisation states and spatial modes~\cite{Hufnagel:2019} (see Figure~\ref{fig:underwater-experiment}). 
A Shack-Hartmann wavefront sensor~\cite{shack-hartmann} is used to record the distorted beam’s wavefront and to analyse the turbulence of the underwater channel using Zernike coefficients estimated in real-time.
The same paper explores the feasibility of transmitting polarisation states as well as spatial modes through the underwater channel for applications in quantum cryptography.

\begin{figure}[htp]
    \centering
    \includegraphics[width=0.49\textwidth]{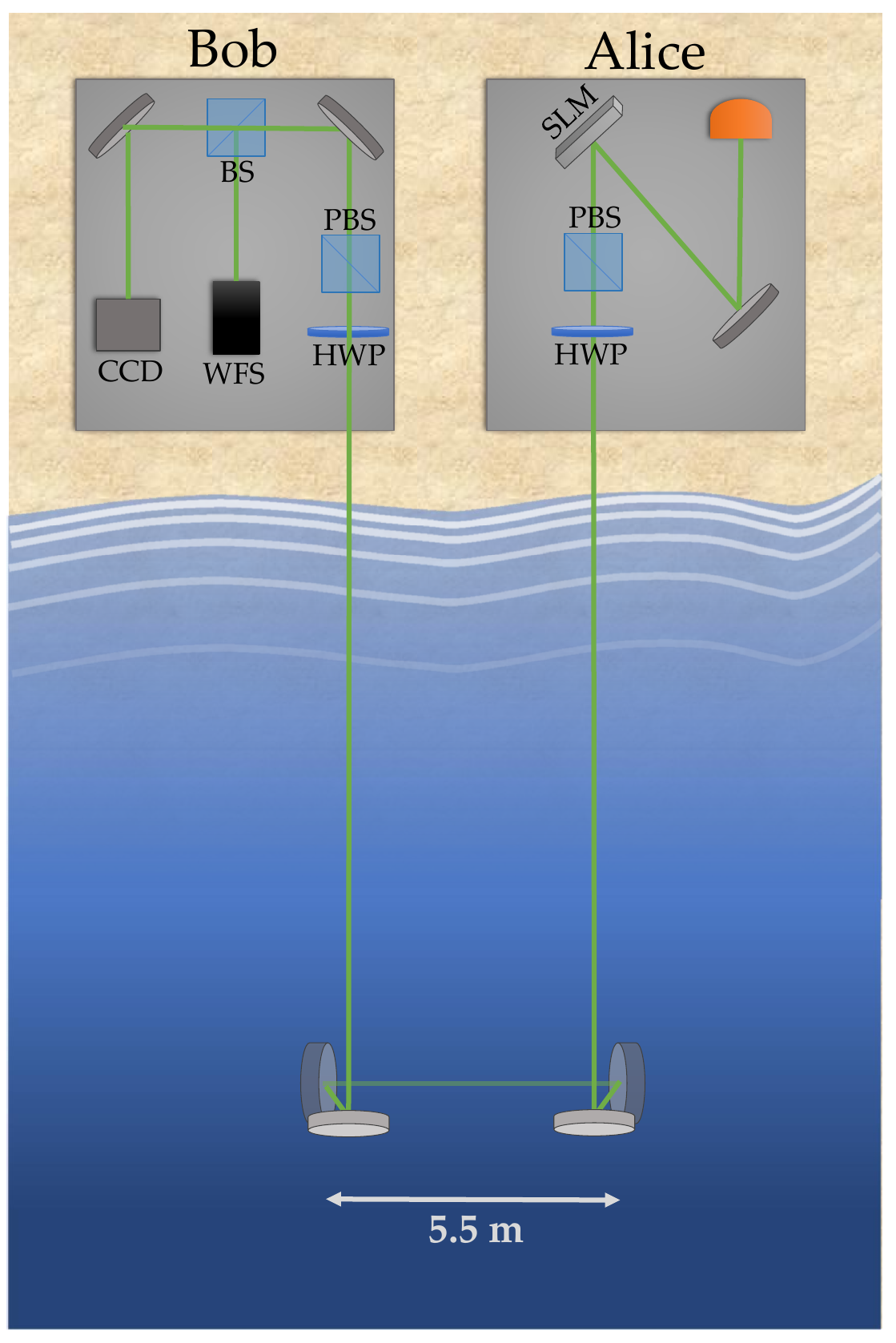}
    \caption{ 
Alice and Bob are the sender and receiver, respectively, in an experiment that involves two breadboards placed on a beach. To prepare the state, a CW laser at $\lambda=532$ nm is sent to an \gls{SLM}, Polarising Beamsplitter (PBS), and Half-Wave Plate (HWP) at Alice's end. The beam is then directed to a first periscope, which brings it underwater, and it propagates to the second periscope 5.5 m away. At the receiver's end, there is a PBS and HWP for polarisation measurements, and a beam splitter allows a Charged-Coupled Device (CCD) camera and Shack-Hartmann Wavefront Sensor (WFS) to take images.
    Adapted from~\cite{Hufnagel:2019}.}
    \label{fig:underwater-experiment}
\end{figure}

The recent development of network-based underwater navigation systems~\cite{Munafo:2017jfr}, makes quantum communication interesting from a navigational perspective as well.
The ability to secure communication, paves the way to add a level of security when navigational data is exchanged, and hence makes it possible for the communicating nodes not to give away their positions.
Although underwater quantum optical communication is still in its early phases, as the technology matures the ability to resolve quantum features could lead to direct improvements in localisation, at least at short ranges.
For example, \cite{Lanzagorta:2014} discussed application of quantum sensing modalities for detection of underwater vehicles in very shallow waters (i.e., less than \SI{20}{\meter}), outperforming results achieved using radar, LIDAR or magnetic anomaly detection~\cite{DBLP:series/synthesis/2006Holmes}. 
\\

\subsection{Quantum key distribution (QKD)}
\label{sec:qkd}
The most mature application of quantum communication is ~\gls{QKD}, a technique used in quantum cryptography which consists in generating and distributing a key between two or more parties for encrypt and decrypt messages, then distributed through classic channels. 
is easily experimentally realised.  
\gls{QKD} has made it possible to demonstrate that the quantum properties of entangled states survive even after penetrating the atmosphere in long-range quatum communication tests. 

An important aspect of \gls{QKD} is that it guarantees a - mathematically proven - security that classical systems cannot achieve. Any third party trying to gain knowledge of the key will inevitably disturb the states of the transmitted qubits showing its presence (for a more in-depth description and mathematical proof of the protocol's security see 
\cite{nielsen_chuang_2010}).
There are several \gls{QKD} protocols, which are distinguished by qubit source, number of qubits transmitted, base used for the measurement of states and so on. Below we will describe the general scheme of the so-called BB84 protocol, the first protocol for quantum cryptography.\\

Let's imagine that Alice wants to share a secret message with Bob. 
First of all Alice chooses a string $S$ of classical bits - 0 and 1 - and randomly encode each data bit in one of her two bases: $\{|0\rangle, |1\rangle\}$ if the corresponding bit is 0, and $\{|+\rangle, |-\rangle\}$ if the corresponding bit is 1.  Alice then sends the resulting states to Bob, using the quantum channel and Bob measures each received qubit in the $\{|0\rangle, |1\rangle\}$ or $\{|+\rangle, |-\rangle\}$ basis as random. Now the fundamental point is that, since the photons are entangled, each pair will have opposite polarization, so when Bob measure its qubit with the same basis used by Alice to encrypt the original bit, he will get the same sequence $S$ of 0 and 1. At this point, the last step to do will be to communicate, through a classic channel, the basis used by Alice and Bob and discard the values corresponding to operations made with different bases. The remaining sequence, which is not broadcast via the channel, will be the key to encrypt and decrypt their messages.
The strength of this process is its security, as if an eavesdropper tries to intercept the message he must inevitably interact with the photons changing their state and creating inconsistencies between Alice and Bob's keys. In Figure~\ref{fig:qkd} an example of a key measurement and sharing process between Alice and Bob is shown.

\begin{figure}[htp]
\centering
\includegraphics[width=0.5\textwidth]{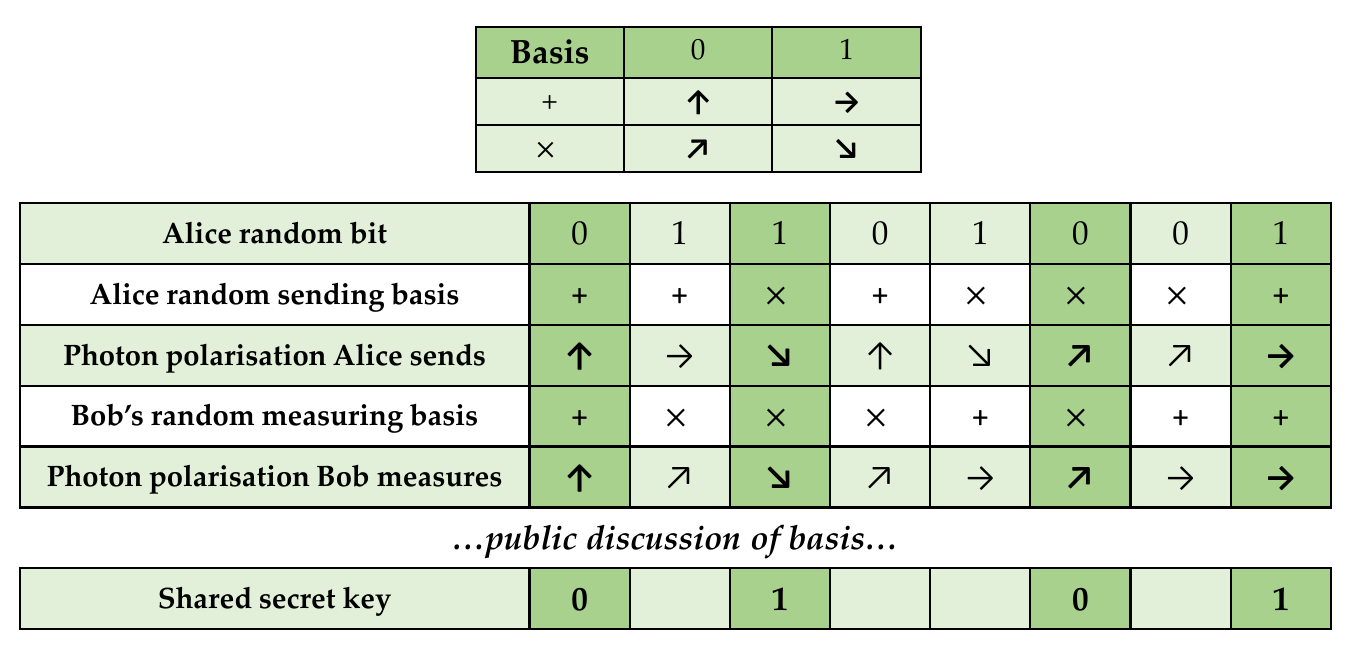}%
\caption{A simple example of generating a key using \gls{QKD}. Adapted from~\cite{qkd_image}.}
\label{fig:qkd}
\end{figure}

The \gls{QKD} process is a simple test that shows the potential of some features of quantum mechanics such as the uncertainty principle, quantum entanglement and measurement on a quantum system.

From a navigation perspective, the ability to improve communication security is important to ensure that navigation aids and the relevant communications come only from trusted nodes, requiring confidentiality/integrity/authentication.
High level of security means that malicious actors cannot impair, mislead or hijack \glspl{AUV}, and hence reduce the ability of the vehicles to successfully perform a mission.

\subsection{Quantum clocks and clock synchronisation}
\label{sec:quantum-clock-synchronisation}

Clock synchronisation is key to a number of applications, including satellite navigation (see also Section~\ref{sec:quantum:positioning-systems}), network-based navigation, communication systems, and for 
coherent distributed-aperture sensing at high frequencies.
Atomic clocks, from what is known as the first quantum revolution, are based on atomic microwave frequency standards (the second was officially redefined by the Comité International des Poids et Mesures in terms of the gap between two specific energy levels in a caesium-133 atom in 1967) that are used to generate accurate and precise time and frequency with short-term instability better than $\approx1\times10^{-13}\;at\;1s$ using quartz oscillators and of $\approx1\times10^{-15}\;at\;1s$ 
using optical transitions (i.e., using light rather than microwaves)~\cite{https://doi.org/10.48550/arxiv.2004.09987}.

Although atomic clocks are the most mature technology to date and represent the standards for precision time and frequency metrology, it has been shown that by making greater use of the quantum properties of atoms it is possible to obtain higher stability and lower uncertainty. 
Optical clocks, which are atomic clocks based on optical energy-level transitions instead of the more traditional microwaves transitions, go in that direction. 
The first optical clock to achieve lower levels of uncertainty than caesium clocks was the Quantum - Logic Clock~\cite{QLogicClock}: an optical atomic clock based on quantum-logic spectroscopy, a tool that has the purpose of reading the electronic state of ions that do not have transitions suitable to be detected - spectroscopic ions - by transferring its state to another ion - called logical ion - trapped in the same potential. Quantum - Logic Clock exploits a single trapped $^{27} {\mathrm{Al}}^{+}$ ion transition and has an
oscillation uncertainty is below $ 10^{-18}$. 
The second type of optical clocks considered involve neutral atoms trapped in an optical lattice. 
An optical lattice is nothing more than a series of laser beams arranged in such a way to interfere and create a standing wave. This can extend in two or three directions, generating respectively an \textit{egg carton - like} lattice, or even a 3D grid, trapping atoms or molecules in the peaks and valleys of the standing wave. In 2015 JILA (Joint Institute for Laboratory Astrophysics) improved the stability of their $^{87}\mathrm{Sr}$ optical lattice clock reaching an absolute frequency uncertainty of $2.1\times 10^{-18}$~\cite{Nicholson2015}. Two years later in the same institute a 3D quantum gas optical lattice clock was realised and to date it has achieved an accuracy of $2.5\times 10^{-19}$ over 6 hours.


Quantum clocks hold the promise to achieve much higher time measurement precision~\cite{krelina2021quantum}, and over the last ten years, a number of clock synchronisation schemes have been proposed based on quantum mechanical concepts. 

Early work on quantum-based clock synchronisation~\cite{PhysRevLett.85.2006} leveraged quantum properties to determine the time difference $\Delta$ between two spatially separated clocks minimising the number of messages that needs to be exchanged between the clocks. The algorithm, named "ticking qubit", relies on transmitting qubits ticking at different frequencies~\cite{Bennett2000QuantumIA}.
The qubits hence behave naturally like a small clock during their transit, and this makes it possible to reduce the number of messages (i.e., qubits) that need to be exchanged.
The number of frequencies used is key to reducing the number of qubits needed. \cite{https://doi.org/10.48550/arxiv.cs/0103021} presents an optimal algorithm that is able to time synchronise to $n$ bits of accuracy while communicating only one qubit in one direction using an $O(2^n)$ range of frequencies.

Usage of shared prior quantum entanglement and classical communications makes it possible to synchronise a pair of atomic clocks that are spatially separated without the need for the two parties of knowing their relative location or the properties of the intervening medium~\cite{PhysRevLett.85.2010}. 
The approach, which is similar to Ekert’s entanglement-based ~\gls{QKD} protocol~\cite{PhysRevLett.67.661}, has initially only entangled clocks in a global state which does not evolve in time. 
The synchronised clocks are then extracted via measurements and classical communications.
More specifically, the algorithm assumes that Alice (A) and Bob (B) share an ensemble of singlet states (i.e., a system in which all electrons are paired):


\begin{equation}
|\Psi^-\rangle = \frac{1}{\sqrt{2}}\big(|+\rangle_A |- \rangle_B - |-\rangle_A |+ \rangle_B \big)
\end{equation}

where $|\Psi^-\rangle $ is a state that does not evolve in time, and for this reason all the pre-clock pairs are "idling", i.e., they can provide no direct timing information, while $|+\rangle$ and $|-\rangle$ are defined in (\ref{eq:hadamard+}) and (\ref{eq:hadamard-}).
To start the clocks each party measures all of her pre-clock pairs, and when they do so, they collapse, with equal probability, randomly and simultaneously at both A and B, into one of two states:


\begin{equation}
\begin{split}
&|\Psi^{I}\rangle = |+\rangle_A |- \rangle_B \\
&|\Psi^{II}\rangle =|-\rangle_A |+ \rangle_B
\end{split}
\end{equation}

When the clocks start to evolve in time, Alice can measure its clock, and to complete the clock synchronisation she needs to communicate to Bob the relevant information (i.e., labels to specify which subset of Bob's qubits are of the same type she has) using classical communications. Using this information Bob can then measure his own subset of qubits to finally have synchronised clocks.

It is interesting to note that actual synchronisation does not require any exchange of information containing timing information and it has no restriction on the mode of communication or the properties of the intervening medium.

This approach is particularly attractive for satellite constellations because the space-borne atomic clocks' ability to synchronise with a master atomic clock on the ground is affected by the fluctuating refractive index of the atmosphere, which causes random variations in the speed of light and limits the accuracy of the traditional approach.

The protocol can be generalised to a multi-party version~\cite{PhysRevA.66.024305} supplying the $n$ parties with shared $n$-qubit systems in known entangled energy eigenstates and then applying the same approach discussed in~\cite{PhysRevLett.67.661}.
\cite{Kong:2018} implemented the protocol experimentally, demonstrating very good agreements between theoretical predictions and experimental results.

\cite{giovannetti2001} derived quantum clock synchronisation based on \gls{TOA} and demonstrating that quantum entanglement and squeezing~\cite{fox2006quantum} make it possible to increase the accuracy of the synchronisation procedure by a factor $\sqrt{MN}$, using $M$ pulses of $N$ photons each, as compared to positioning using unentangled and unqueezed pulses with the same bandwidth. 
The time synchronisation algorithm is obtained as a byproduct of the more general problem of localising the position of $n$ receivers using \gls{TOA}.
To perform clock synchronisation M light pulses are transmitted to M detectors, measuring and then averaging the corresponding \gls{TOA}. The accuracy of the method depends on the number of pulses and the number of photons per pulse.
More details on this method are reported in Section~\ref{sec:quantum:positioning-systems}.


A protocol for synchronising distant clocks that does not rely on the arrival times of signals is proposed in~\cite{giovannetti2004}. 
This method, called the conveyor belt synchronisation scheme and shown in Figure~\ref{fig:conveyor-belt} is based on encoding synchronisation information on the pulses using the polarisation direction, frequency, or phase.
The algorithm works by having two parties, Alice and Bob, interact with a conveyor belt in a specific way. Alice pours a quantity of sand proportional to the time shown on her clock on both sides of the conveyor belt. Bob, located in between Alice's pouring points, scoops away a quantity of sand proportional to twice the time shown on his clock.
If the proportionality constants for Alice and Bob are the same, then the quantity of sand at a specific point D on the conveyor belt will be constant in time, after an initial transient period when they begin to act on the system. The time difference between the two clocks can be determined by measuring the amount of sand at point D.
This method is not affected by any dispersion that may be present in the medium through which the light signals are exchanged, and the use of frequency-entangled pulses offers greater immunity to dispersion than the classical version of the protocol.

\begin{figure}[htp]
    \centering
    \includegraphics[width=0.5\textwidth]{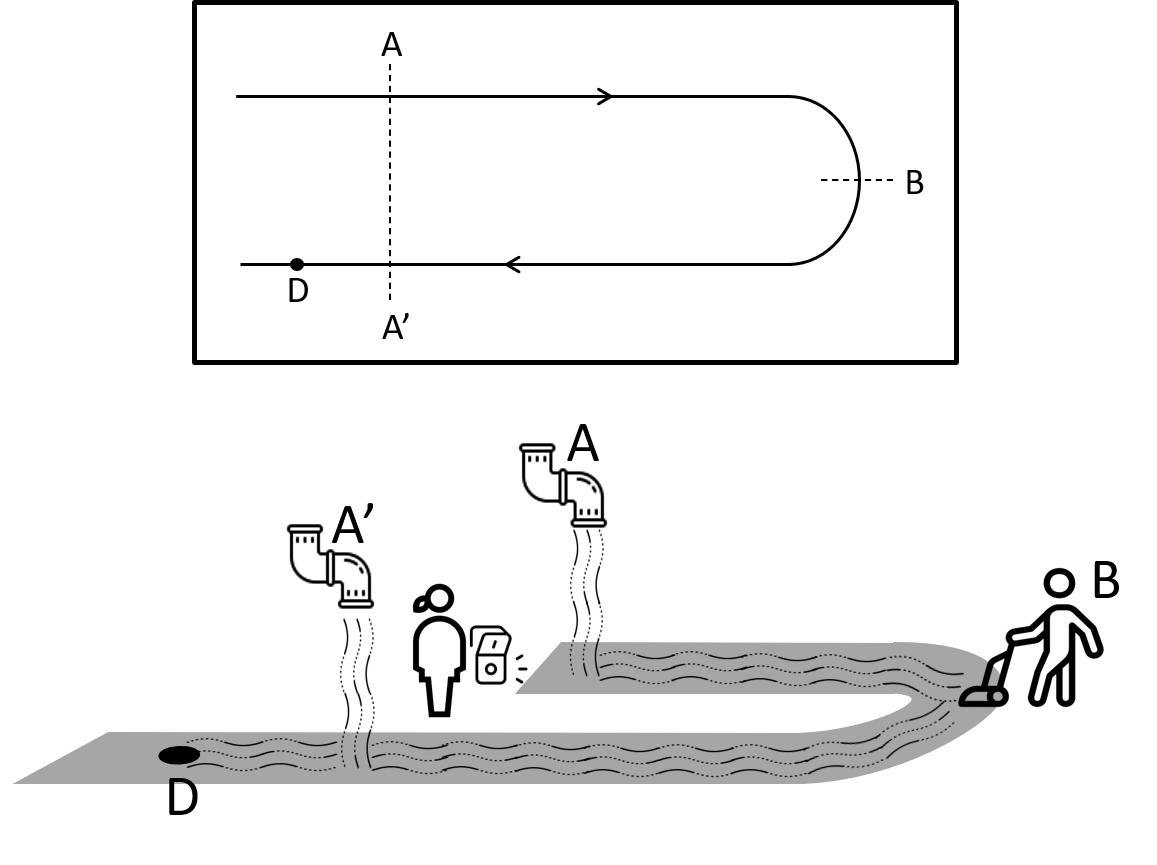}
    \caption{A visual illustration of the conveyor belt synchronization method is shown where Alice adds sand on the belt at points A and A', and Bob removes sand at the middle point B. The difference in their clocks can be determined by measuring the amount of sand at position D after the initial transient period. Adapted from~\cite{giovannetti2004}.}
    \label{fig:conveyor-belt}
\end{figure}

Clock synchronisation is at the centre of applications such as high-accuracy satellite navigation or geolocation.
It is also makes it possible to use \gls{OWTT} for localisation.
According to this method, a reference system (e.g., a surface vehicle), periodically transmits an acoustic data packet containing its accurate position (for example because it has access to \gls{GPS}) 
together with the time the data packet was transmitted. 
Having an accurate time base, the receiver can compare the time it received the message with the time it was transmitted to calculate its distance from the reference system.
Although, existing \gls{CSAC} are stable enough to enable short duration \gls{OWTT} navigation, their current drift rate is too big for long-duration missions (see also Section~\ref{sec:discussions}). 
Quantum-clocks can overcome some of the existing limitations improving time resolution and reducing clock drift substantially.
More broadly, the recent experimental results obtained in quantum clock synchronisation have boosted interest in developing a wide variety of algorithms some of which are specifically designed for localisation and navigation.
The reader is referred to~\cite{QPSsurvey} for a survey on clock synchronisation with specific focus on positioning systems.

\subsection{Quantum imaging}
\label{sec:quantum-imaging}
Quantum imaging is a sub-field of quantum optics that exploits the quantum entanglement of the electromagnetic field, and photon correlations effects, to image objects with characteristics that go beyond what is possible in classical optics~\cite{Lugiato_2002}. This makes it possible to improve image resolution, \gls{SNR}, contrast, and spectral range, allowing for spectroscopy and imaging in spectral ranges where no efficient detection is currently possible or even imaging with light that actually never interacted with a sample, or at extreme low light intensity.
Early experiments dated back to the 1970s where the temporal and spatial correlations between pairs of photons were investigated~\cite{PhysRevLett.25.84}.
An interesting application of quantum imaging is quantum ghost imaging which uses optical correlations to enable image acquisition even when information from one of the detectors used during the acquisition does not yield an image (i.e., out of the line of sight of the camera)~\cite{Moreau:2017}.

In a ghost imaging system, shown in Figure~\ref{fig:quantum-imaging}, the main detector is a single-element (single-pixel) device that measures the interaction between a single photon and the unknown object. 
The position of the other, spatially correlated photon is measured using an imaging system without interacting with the object.
A ghost imaging system sources two entangled photons of different - or equal - frequencies. 
The photon at the optical frequency is recorded directly by a high-resolution photon-counting camera while the other photon at a different frequency (e.g. infrared) is sent towards the object. 
The reflected photon is then detected by a single-photon detector, known as a "bucket" detector. 
The image is created by analysing the correlations between both photons. 
This means that an object can be illuminated at one wavelength, while the spatial information is recorded at another wavelength where the imaging detector is more sensitive or less noisy.

This schema makes it possible to image objects at extremely low light levels, and it has been demonstrated in cases where infrared light can penetrate with a better \gls{SNR}~\cite{WALBORN201087}. 
These properties of quantum imaging make it particularly attractive for underwater navigation where quantum sensor can have 5x increased performance in the low signal-to-noise ratio and low brightness regime when compared with classical sensors ~\cite{10.1117/12.2262654}, with the additional advantages of improving the operational stealthiness of the system thanks to its ability to retain high performance within well defined and narrow bandwidths. 
Results are mostly theoretical with many challenges that must be overcome to make the system of any operational relevance, but there seems to be general consensus that quantum imaging for underwater navigation deserves further scientific and engineering consideration, research, and discussion~\cite{10.1117/12.2262654}.
Additionally, the ability to improve image resolution and to illuminate objects at different wavelengths through ghost imaging might make it possible to obtain a larger number of image features that can then be used to create better optical maps for \gls{SLAM} or visual-based navigation.

\begin{figure}[htp]
    \centering
    \includegraphics[width=0.5\textwidth]{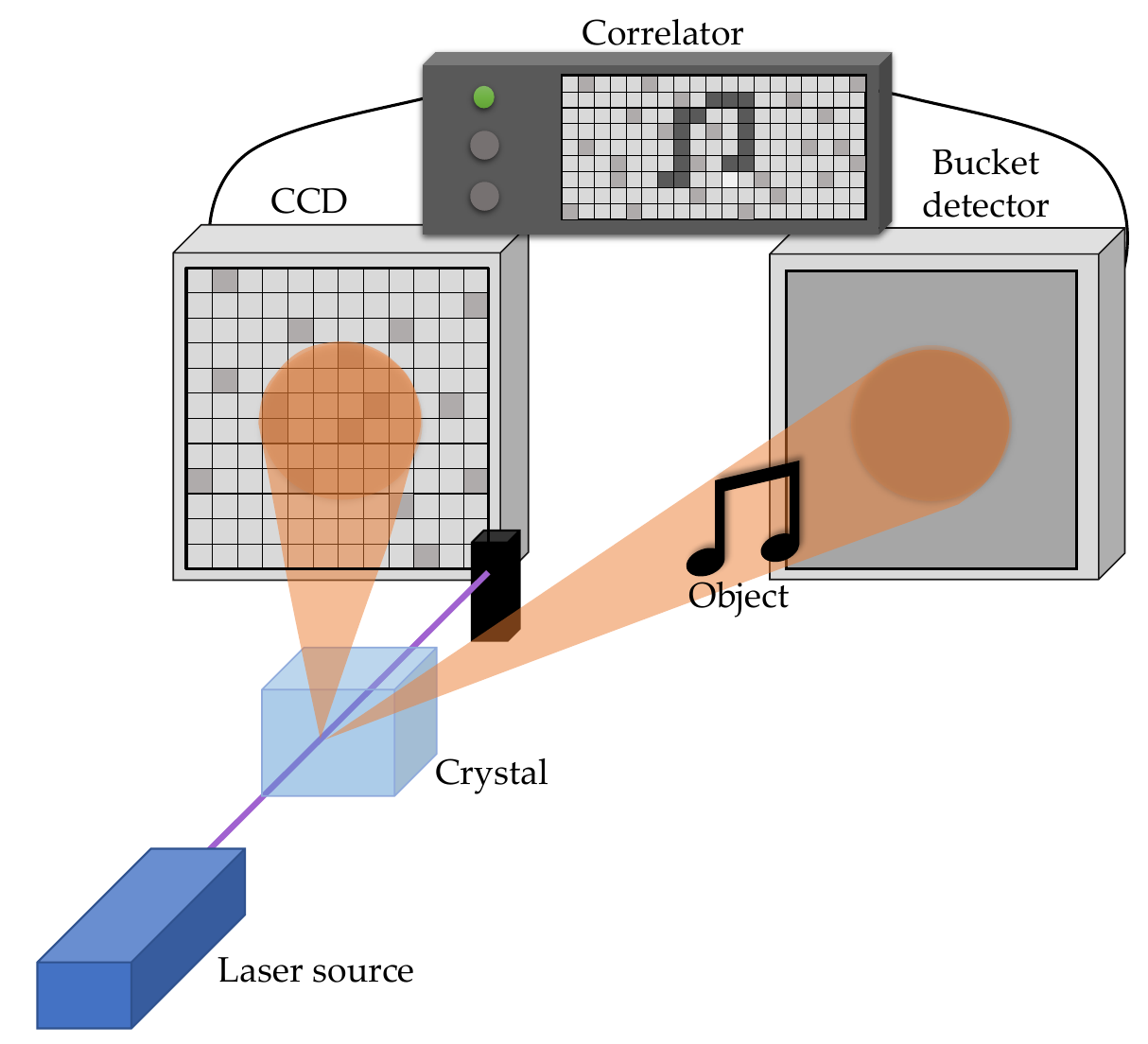}\\
    \caption{Ghost imaging scheme. A laser beam is sent towards a nonlinear crystal. The two beams resulting from this interaction (see Appendix~\ref{sec:quantum-entangled-states})  are correlated and one is sent towards a detector with spatial resolution, the other hits the object and is detected by a single pixel camera (Bucket detector). The image is created by correlating information coming from both detectors.}
    \label{fig:quantum-imaging}
\end{figure}

\subsection{Quantum Positioning Systems}
\label{sec:quantum:positioning-systems}

The \gls{GPS} system consists of approximately 24 satellites orbiting Earth at around 4.25 Earth radii. To calculate unknown user space-time coordinates $(t_o, r_o)$, four equations ~\cite{GPS-Book} are solved using known satellite coordinates $(t_s, r_s)$ and assuming line-of-sight signal propagation. 
\gls{GPS} signals are transmitted in the L-band with frequencies $L1 \approx 1575.42$ MHz and $L2 \approx 1227.6$ MHz. 
Pseudorange measurements are obtained by comparing the \gls{PRN} code received from the satellite and the replicated \gls{PRN} code in the \gls{GPS} receiver. 
Clock synchronisation is crucial for the system, and its precision depends on the arrival time measurement of light pulses~\cite{GPS-Book}. 

Quantum entanglement and squeezing 
can be used to overcome the classical limitations of conventional techniques, resulting in improved accuracy and performance in timing and localisation. 
By utilising quantum mechanics, the accuracy of determining the arrival time of a light pulse is dependent on the bandwidth of the pulse and the number of photons or power in the pulse. 
This means that by increasing the number of photons in a quantum pulse, the accuracy can be enhanced. 
For example, using 100 photons results in a tenfold enhancement over the classical limit, and using 1 million photons results in a 1000-fold improvement.

A discussion on how quantum entanglement and squeezing can be used to improve the accuracy of positioning systems, clock synchronisation, and ranging is reported in ~\cite{giovannetti2001}. 
The paper reports that by using \gls{TOA}-based methods, the accuracy of the position, which depends on the number of pulses $M$, the bandwidth of the pulses, and the number of photons per pulse, can be increased by $\sqrt{M}$ when compared to positioning using unentangled pulses with the same bandwidth.

Using classical methods, the \gls{TOA} of each pulse has an intrinsic indeterminacy that depends on the spectrical characteristics and mean number of photons $N$ of the pulse.
For example, given a gaussian pulse of frequency spread $\Delta\omega$, the \gls{TOA} estimated using $N$ data points cannot be measured with an accuracy better than $\frac{1}{\Delta\omega\sqrt{N}}$ (note that this corresponds to the fact that the times of arrival of the single photons has an indeterminacy $\frac{1}{\Delta\omega}$).
Employing frequency-entangled pulses gives an increase in accuracy by a factor $\sqrt{M}$ in the measurement of time with respect to the case of unentangled photons.
Employing number-squeezed states makes it possible to enhance positioning:~\cite{giovannetti2001} shows that the $N$-photon Fock state gives an accuracy increase of $\sqrt{N}$ compared to the coherent state with mean number of photons $N$ and that entangled pulses of number-squeezed states combine both these enhancements, hence providing an overall improvement of $\sqrt{MN}$ in the mean time of arrival, over the accuracy obtainable by using $M$ classical pulses of $N$ photons each.

Finally, using the same reasoning and assuming that the distance between the participating parties are known, the authors also show that clock synchronisation can be significantly enhanced as compared to classical protocols using light of the same frequency and power. 
More details on clock-synchronisation is reported in Section~\ref{sec:quantum-clock-synchronisation}.

The \gls{QPS} method discussed in \cite{giovannetti2001} is sensitive to loss. If one or more of the photons fails to arrive, the time of arrival of the remaining photons does not convey any timing information. 
Possible workarounds include ignoring all trials where one or more photons is lost, or the usage of partially entangled states, which provide a lower level of accuracy than fully entangled states, but are more tolerant to loss~\cite{PhysRevA.65.022309}.
Moreover, preparing lots of photons in the requisite state is hard and requires precise application of nonlinear optics and photonics. For $M=2$, a continuous-wave parametric down-converter (see Appendix~\ref{sec:quantum-entangled-states}) can be used to output the required twin beam state.
However, for $M > 2$, the creation of such frequency-entangled states represents a continuous variable generalization of the Greenberger–Horne–Zeilinger state, and, as such, is a considerable experimental challenge.

Building on these results, and on the quantum algorithms proposed to improve the clock synchronisation between satellites and receivers (see among others~\cite{giovannetti2001, PhysRevLett.85.2006, PhysRevLett.85.2010, PhysRevLett.87.117902, PhysRevA.65.022309, Bahder_2004}),~\cite{bahder:2004} proposes an algorithm to directly determine all four space-time coordinates using \gls{HOM} interferometery.

To calculate all four space-time coordinates, the \gls{QPS} can be based on entangled photon pairs (biphotons) and second order correlations within each pair. 

\subsubsection{Earth-based and space-based systems}
The space quantum communication scheme can be divided into two categories: earth-based and space-based. 
In the earth-based scheme, a ground-based transmitter is used to distribute single photons or entangled photon pairs to ground stations and satellites, enabling quantum teleportation of particle states between different terminals.
This method has a limited communication distance, as the link is through land-free space. 
On the other hand, the space-based scheme utilises a light source on a space transmitting platform, and the optical communication link is established through an intersatellite channel. 
This method is less affected by atmospheric turbulence, resulting in longer communication links and a more global quantum teleportation scheme that is easier to achieve.

The two-photon coincidence counting rate is the basic measured quantity. 
In order to determine his four space-time coordinates, a user of the \gls{QPS}  must carry a corner cube reflector, a good clock, and have a two-way classical channel for communication with the origin of the reference frame.
In this case, the \gls{QPS} becomes the quantum equivalent of the classical~\gls{GPS} (see Figure~\ref{fig:bahder}).
However, while in a classical \gls{TOA} system like the \gls{GPS}, receivers and transmitters must have clocks that are synchronised with a long-term stability, in \gls{QPS}, clocks only require short-term stability to update a photon coincidence counter~\cite{bahder:2004}.
More broadly, time synchronisation can be achieved using a clock synchronisation algorithm that can itself rely on quantum entanglement~\cite{Bahder:2004b}.
The determination of the position is based on a controllable optical delay located at one of the photon detectors of the baseline.
When photons go through path from the baseline to the user position and back they follow the same path apart from those going through the optical delay. 
The presence of the optical delay is used to obtain a balanced interferometer which corresponds to the case where the round trip time of the photons is the same regardless of the traversed path. 
A balanced interferometer provides a unique minimum in the two-photon counting rate, which can be measured with an accuracy that depends on the bandwidth $\Delta\omega$ of the band-pass interference filters used in front of the photon detectors.
Once all interferometers in the baseline are balanced, it is then possible to relate the geometric path lengths to the measured optical delay time for each of the baseline elements
using classical \gls{TOA} system that records arrival times of classical light pulses,
and obtain a system of equations that can be solved to obtain the spatial coordinates of the reflector.
The method makes it possible to calculate the position of the reflector using quantum counting that ensures a unique observable minimum in the two-photon coincidence counting, with the search for the optical delays to obtain balanced interferometers that corresponds to the quantum equivalent of the correlation in classical \gls{GPS}.
Classical communications are needed to share the information that established when all interferometers are balanced and that hence the system of equations can be solved. The author points out that a similar method could be applied using lasers and applying classical physics, but that in this case the interference fringes might create ambiguities that cannot be easily resolved.
Since the method relied on correlations between photon pairs, the number of baselines needed is three (versus four that is needed for classical \gls{GPS}).
More fundamentally, and more significant for applications, is that in the classical case of a \gls{TOA} system, clocks must be synchronised to coordinate time so that accurate pulse arrival times at the four reception points can be recorded.
Again, using two-photon coincidence counts makes it possible to use short-term stability
clocks that guarantee synchronisation while the optical time delay is adjusted.
Results are provided in terms of \gls{GDOP}(i.e., the effect of geometrical positions of the GPS satellites on the
accuracy of the user’s position), and the author suggests that a satellite-based \gls{QPS} may achieve a position accuracy near the Earth's surface of \SI{1}{\cm}, while recognising that significant engineering is required to implement the proposed scheme in practise.

\begin{figure}[htp]
    \centering
    \includegraphics[width=0.5\textwidth]{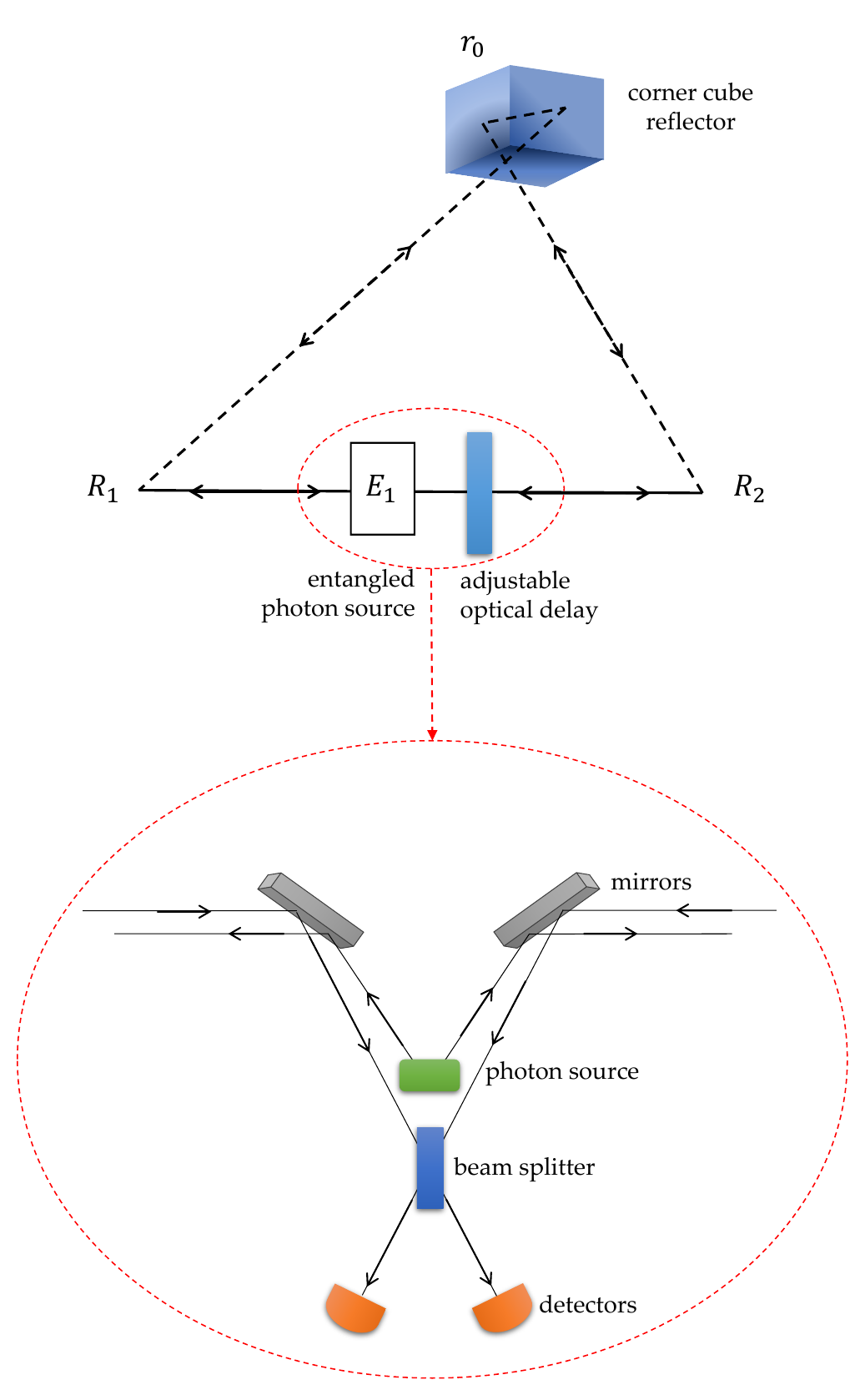}
    \caption{Set-up, proposed in \cite{bahder:2004}, to calculate all four space-time coordinates: a user of the \gls{QPS} must carry
a corner cube reflector, a good clock, and have a two-way classical channel for communication.}
    \label{fig:bahder}
\end{figure}

\subsubsection{RF-photonics sensing and distributed quantum sensing}
Using {RF-photonics} sensing to convert \gls{RF} waves into the optical domain and then using quantum entanglement to increase the photons' sensing capabilities, ~\cite{PhysRevLett.124.150502} demonstrated experimentally that they can boost the level of precision obtainable in sensor networks using distributed quantum sensing, effectively using quantum correlation between multiple sensors to enhance the measurement of unknown parameters beyond the limits of unentangled systems~\cite{PhysRevA.97.032329}.
The experiment demonstrated for the first time that a network of three sensors can
be entangled with one another, meaning they all receive the information from probes and
correlate it with one another simultaneously.

\subsection{Quantum acoustics}

The previous sections focused mostly on quantum optics, which deals with how photons, individual quanta of light, interact with atoms and molecules. 
Quantum technologies can be used for ultra-precise sound sensing up to the level of a phonon, a quasiparticle quantising sound waves in solid matter~\cite{doi:10.1126/science.aao1511, Satzinger:Nature:2018}, using photoacoustic detection: measurement and study of sound waves inducted by light. 
Quantum mechanics applies equally to macroscropic mechanical systems, and this has been experimentally demonstrated in the last decade in two main ways: leveraging cavity optomechanics, where the position of a mechanical oscillator parametrically couples to a higher-frequency electromagnetic cavity~\cite{RevModPhys.86.1391}, and via quantum acoustics, where an artificial atom or qubit exchanges quanta with a mechanical oscillator~\cite{Connell:2010}. 
{Quantum acoustics is the acoustic analogue of cavity or circuit quantum electrodynamics (cQED)}.
Quantum acoustics studies phonons, packets of vibrational energy emitted by jittery atoms, which manifest as sound or heat, depending on their frequencies.
Phonons are, like photons (see Section~\ref{sec:a-primer-to-quantum-physical-functioning}), restricted to have discrete values of vibrational energy.

Revealing the quantised structure of energy in a system is only possible with apparatuses that have a precision greater than the energy of a single phonon. 
One of the first experiments that achieved this was reported in~\cite{Arriola:Nature:2019}, where an artificial atom was used to sense the motional energy of a driven nanomechanical oscillator with sufficient sensitivity to resolve the quantisation of its energy. 
The experiment involved exciting phonons with resonant pulses and probing the resulting excitation spectrum of the qubit to observe phonon-number-dependent frequency shifts that were about five times larger than the qubit coherence time. 
The results demonstrated an integrated platform for quantum acoustics that combines large couplings, considerable coherence times and excellent control over the mechanical mode structure (see Figure~\ref{fig:resolving-energy-levels-of-nanomechanical-oscillators}).

\begin{figure}[htp]
    \centering
    \includegraphics[width=0.5\textwidth]{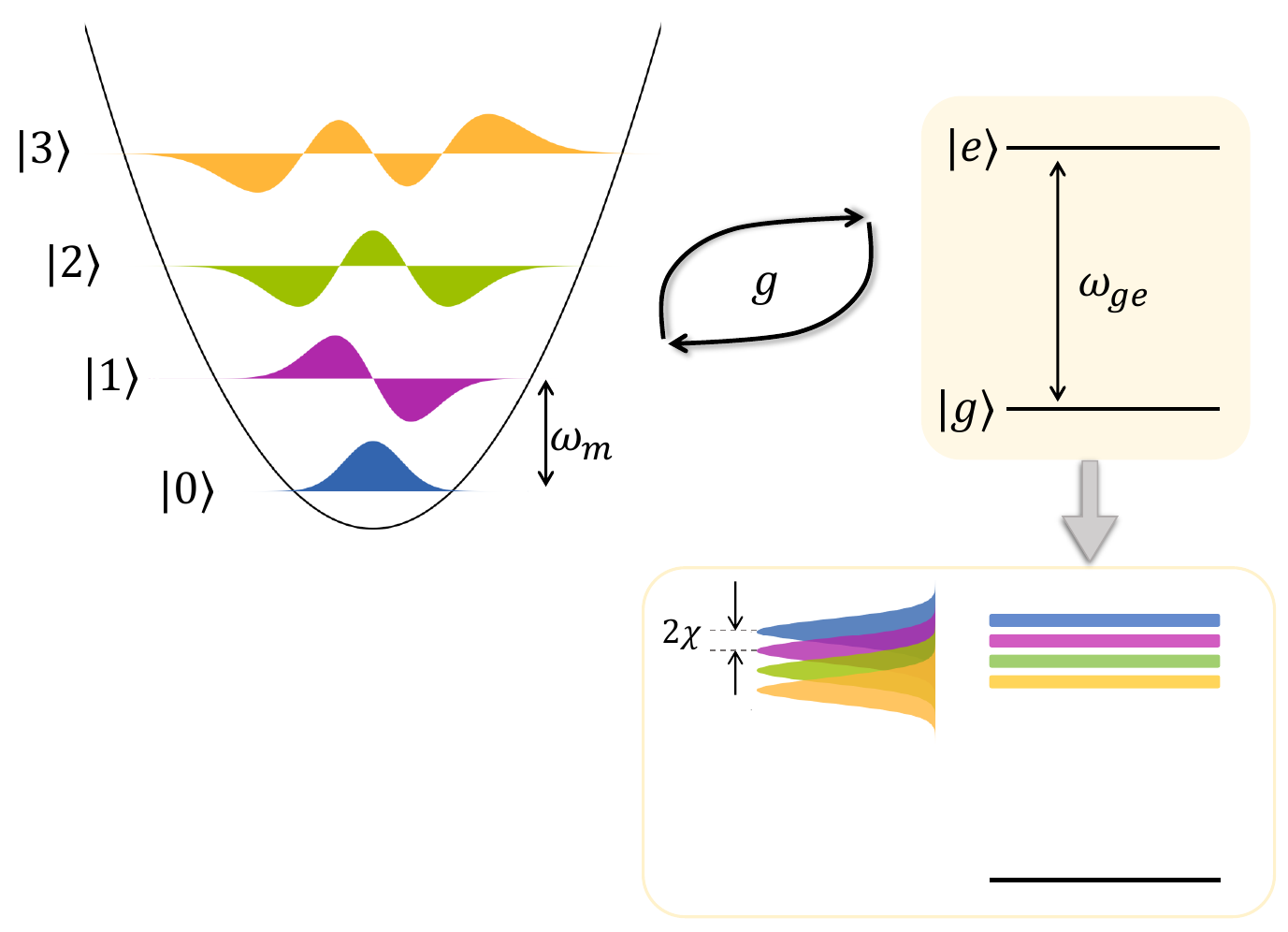}
    \caption{In quantum mechanics, a mechanical oscillator's state is described by a linear superposition of energy eigenstates $|n\rangle$, each representing a state of n phonons. These energy levels are equally spaced and cannot be distinguished because they all have the same transition frequency $\omega_m$. However, by coupling the resonator to a qubit with a transition frequency $\omega_{ge}$ at a rate of $g$, the qubit spectrum splits. This is parameterised by the dispersive coupling rate $\chi$. This allows the identification of the different phonon-number states present in the oscillator.
    Adapted from~\cite{Arriola:Nature:2019}.}
    \label{fig:resolving-energy-levels-of-nanomechanical-oscillators}
\end{figure}

An effective approach for quantum acoustics is to connect a superconducting qubit to a \gls{SAW} resonator and place them on different substrates to attain a better \gls{SNR} and regulate the coupling strength of the components. 
This approach enables quantum experiments to verify that the phonons within the \gls{SAW} resonator are in quantum Fock states by means of quantum tomography~\cite{Satzinger:Nature:2018}. 
Similar techniques have been attempted by utilising bulk acoustic resonators~\cite{Chu:Nature:2018}. 
As a result of these advancements, it is now possible to study the properties of atoms that are much larger than conventionally found by modeling them using a superconducting qubit coupled with a \gls{SAW} resonator~\cite{Andersson:Nature:2019}.%

The XARION Laser Acoustics Company is developing a new type of acoustic sensors whose technology is based on detecting acoustic pressure waves optically via a miniature interferometric cavity formed by two parallel millimetre-size mirrors that is able to sense small change in the refractive index of the sound-propagating medium within the cavity itself~\cite{Satzinger:Nature:2018}.
The size of the system does not scale with the sensitivity, as for conventional acoustic microphones and it has a very flat frequency response, from about \SI{5}{\hertz} (where laser drift starts to dominate) to frequencies of up to \SI{1}{\mega\hertz} (in air).
When used in water, the transducer is operable up to frequencies of \SI{50}{\mega\hertz}, after which point the sound wave approaches the dimension of the laser beam diameter and the transducer is not able to generate an output signal.
Current usage is focusing on non-destructive testing and acoustic process monitoring, leveraging the flat ultra sound response to extract useful information, but it is possible to see applications to underwater acoustics, where new types of transducers could be designed with a flatter response and with the ability to detect higher frequency signals in lower \gls{SNR} or reverberant environments.

Precise detection of acoustic waves can be a key enabler for step changes in many applications, including sonar and navigation~\cite{2017NatComms815331W, Balthasar:2016}.

\section{Discussions}
\label{sec:discussions}

Various quantum technologies are at different \gls{TRL}~\cite{trl-level-doc}, and different authors have different perspectives on their maturity timelines~\cite{uk-roadmap,nato-roadmap,au-roadmap}. 
Quantum-based technology is extremely promising but there appears to be no evident low hanging fruit: most of the results and prototypes are still relatively early and not mature, and some only at an exploration stage (e.g., quantum acoustics).
Figure~\ref{fig:quantum-trl} provides an overview of where this paper sees each of the technology discussed in Section~\ref{sec:sensors-and-technologies-state-of-the-art} and their expected maturity timeline.

Based on our findings, quantum clocks, single photon imaging and point-to-point secure communications might be able to have a relative shorter term \SIrange{0}{5}{year} impact, enabling accurate timing and navigation devices for applications including defence and telecommunications.
As described in Section~\ref{sec:quantum-clock-synchronisation}, timing comes from atomic clocks, either installed directly on vehicles, or leveraging \gls{GNSS} and obtaining time synchronisation with satellite-based atomic clocks.
Existing \gls{AUV} navigation system rely on \gls{CSAC}~\cite{Lutwak2003TheCA, 7761268}.
\gls{COTS} \gls{CSAC} are currently specified with a 10 Part Per Billion (\SI{10}{PPB}) per year aging rate across a wide range of temperatures (\SIrange{10}{70}{\degreeCelsius}).
Note that a clock with a 1 Part Per Billion (\SI{1}{PPB}) error will accumulate 1 nanosecond phase error per second, or roughly \SI{31.5}{\milli\sec} per year~\cite{7761268}.
While these are remarkable results, existing atomic clocks might still limit our ability to correctly measure time for long-term or persistent deployments where vehicles cannot rely on periodic re-synchronisation with external high accuracy clocks (e.g., the \gls{GPS}). 
Recent experimental results for underwater applications are reported in \cite{Kebkal2019} where the authors reported \gls{CSAC} drift of about \SI{2.5}{\micro\sec\per\hour} during \SI{30}{\min} long missions.
In the near future, quantum-based clocks, such as cold-atom or lattice clocks, have the potential to increase our time resolution while keeping the size of the clock small, and to improve the security and resiliency of measuring time, reducing the vulnerabilities to intercepted or blocked satellite signals. This would make it possible to improve the time tracking ability and hence the navigation accuracy of maritime systems in those \gls{GPS}-denied scenarios.
Clock synchronisation between network nodes (including \glspl{AUV}) can allow each node to passively receive broadcasts messages and accurately determine their position relative to the other nodes using \gls{OWTT} range and angle, and hence increasing the scalability of existing network-based localisation systems~\cite{Vermeij:2015, Munafo:2017jfr}.\\

\subsubsection{Communications}
From a communication perspective, one of the more mature field is quantum cryptography (see Section~\ref{sec:underwater-quantum-communications}).
In this case, securing communications means that sharing navigation data among nodes/vehicles is also secure. 
This has a direct impact on enhancing the ability of the vehicles to reliably navigate (see for instance~\cite{Munafo:2017jfr, Paull:2014}).
For example, when external references are used to navigate (e.g., \gls{LBL} or network-based navigation), each node uses the knowledge of the position of known anchor points to localise itself with respect to those points.
Stricter guarantees of data confidentiality, integrity and availability will directly influence the quality of the navigation: depending on the application (e.g., contested environments) adversaries might exploit communication vulnerabilities to maliciously misdirect the vehicle navigation.
The usage of quantum technologies, including \gls{QKD}, will make it possible to achieve level of security that classical cryptography cannot reach.
Quantum technology can provide exponential speed up for vectors of attack on existing asymmetric encryption, and theoretically, on symmetric encryption, and/or for the development of quantum-resilient encryption algorithms~\cite{krelina2021quantum}.
\gls{QKD} results have been demonstrated~\cite{Zhang_2019, india-mod} and commercial application already exists.
However, there exists different perspectives on when and how these quantum methods should be used. 
For example, the \gls{NCSC} of the UK, explicitly discourage the usage of \gls{QKD} for government, military and business-critical applications due to its need for specialised hardware requirements, agreement mechanisms and the requirement for authentication in all use cases~\cite{white-paper-cyber}.
Marine robotics have lagged behind in its usage of secure communications, reliant on the fact that most applications are remote, and on the difficulty to physically get hold of the vehicles.
The typical approach relies on high level of classic cryptography for on-board data.
However, with the development of quantum technology, the risk is that hostile intelligence gathers encrypted data with the expectation of future quantum-based decryption power.
The need to start preparing the vehicles for implementing quantum cryptography is evident both at vehicle level where secrets and confidential data are stored, and at the communication level, where secrets and confidential data are exchanged, with applications for intelligence, military or government. 
Transmitting quantum signals through existing optical fiber networks remains a challenge due to the intense contrast between quantum and conventional signals and the significant loss in networks optical switches.
As a result, the first generation of these devices are likely to function on point-to-point wired links up to 200-300km in length~\cite{Zhang_2019}, with prototypes of point-to-point quantum encryption equipment already available.
In the medium term, secure networks connecting offices or telecommunications switching stations over a larger geographical area might be possible, whereas, in the long term, worldwide quantum communication may be made feasible through fiber optic quantum repeaters or satellite technology.\\

\subsubsection{Quantum Imaging}
Single photon imaging~\cite{Basset_online} are also relatively mature. Experiments have demonstrated that they can be used to perform accurate 3D reconstruction operating in low-light-level conditions and to enlarge the operating range achievable by underwater optical communication systems.
The critical parameters that can limit the system deployability and performance for quantum images are the flux of the single-photon/entangled photon emitter or the single-photon detection resolution and sensitivity, with the requirement of powerful processing. 
\cite{krelina2021quantum} suggests the usage of quantum 3D cameras based on quantum entanglement and photon-number correlations to have unprecedented depth of focus with low noise, and applications to inspections, structural cracks on jets, satellites and other sensitive military technology.
The ability of quantum images to perform at low-\gls{SNR} makes it possible to foresee their applications in challenging underwater environments such as deeper or cloudy waters, extending our ability to leverage new \gls{SLAM} or visual odometry algorithms.
Current \gls{SLAM} algorithms rely on classic cameras or sonars to identify features of interest to simultaneously create a map and localise the vehicle within that map.
Visual odometry can be used to determine odometry information using sequential camera images to estimate the distance travelled and hence the position and orientation of a vehicle.
Their usage underwater however has been so far strongly limited by the available senors: sonar data has limited accuracy, and camera images are strongly limited particularly in low-light and turbid waters, with vehicles that have to travel close to the sea bed or to relevant infrastructures to identify features of interest.
Depending on the mission or on the available vehicle (e.g., hover capable vs torpedo shaped) this might not be possible or it might lead to unsafe vehicle trajectories.
Future low-\gls{SNR} quantum cameras could produce significantly better images at longer ranges, dramatically alleviating existing limitations.
Low-\gls{SNR} quantum imaging could also help in target detection, classification and identification with low signal-to-noise ratios or concealed visible signatures and potentially counter adversaries’ camouflage or other target-deception techniques~\cite{matthews2018}.\\

\subsubsection{Inertial Sensors}
Quantum \glspl{IMU} are expected to have a mid-term impact (e.g., \SIrange{5}{10}{years}), with the expectation to bring exponential improvements when compared to classic \glspl{IMU}.
A prototype quantum-enhanced inertial navigation solution is under active development by \gls{NASA}.
The ability to have precise navigation without external references can have a profound impact for space exploration, for indoor or underground robotics, and certainly for underwater robotics.
At the same time, it is worth highlighting that existing systems are still mostly experimental and/or conceptual, and there are no clear field results yet.
The size and power demands of existing systems are also far from what will be needed for deployments on autonomous robotics systems, including the majority of \glspl{AUV}, but the level of navigational improvements that they promise is such that they represent one of the most interesting sensors. 
Quantum-\gls{INS} is a type of proprioceptive sensor that do not require external infrastructure or active emissions for localisation. 
This makes them ideal for situations where such methods may not be feasible or suitable, including \gls{MCM}, \gls{ASW}~\cite{Hamilton:2020} or marine mammal tracking.
Moreover, the expected advancements in Quantum-\gls{INS} technology are such that future \glspl{AUV} equipped with these sensors will be able to remain underwater for several months without the need to resurface.\\

\subsubsection{Geomagnetic navigation}
Another way to obtain quantum non-\gls{GNSS} long range accurate navigation can be through the usage of quantum magnetometers and gravimeters. 
Geomagnetic navigation has at its core the problem of classifying geomagnetic measurements into corresponding positional classes on a reference map which can be strongly non-linear and where the precision and accuracy of the measurements are key to correctly solve the localisation problem.
Current classical technology does not have the resolution to provide maps of sufficient quality that can be used as navigational references. 
Existing sensors, especially those that are small enough to be mounted on a \gls{AUV} are too limited to detect magnetic and gravity anomalies of significance for navigation.
The expected improvement in performance from both quantum magnetometers and gravimeters, in combination with Earth’s magnetic and gravity anomaly maps, would instead make it possible to estimate the position and velocity of an \gls{AUV} with the necessary precision and accuracy to enable long-range underwater navigation.
Prototypes of quantum magnetometers are commercially available, with some that are of a size that can be supported by \glspl{AUV}~\cite{sbquantum} or that could be towed at the back of the vehicle.
Some commercial companies are starting to look at using magnetometers for navigation but the authors did not find clear results available in the open literature.\\

\subsubsection{Quantum computers and impact on navigation}
Quantum computers are expected to have a significant impact on various fields, although, widespread commercial availability of quantum computers beyond a few narrow, specialised applications will depend on future technological advancements.
They store data using qubits and have been theoretically shown to handle specific types of problems and data exponentially faster than classical computers, as described in Section~\ref{sec:quantum-computing}. 
This can be beneficial in solving non-linear and non-Gaussian state estimation issues that can currently only be approximated.
Sequential Monte Carlo methods, also known as particle filters or recursive Monte Carlo filters, are iterative algorithms that use a set of weighted simulations, or "particles", to approximate a changing target distribution. There are three key steps in particle filter implementation: particle generation, weight calculation, and re-sampling. Re-sampling is particularly important because over time, a small number of weights can become dominant, leading to a poor approximation of the posterior density and inaccurate estimates. To address this issue, particles with large weights are duplicated, and those with negligible weights are eliminated. However, particle filters can be computationally demanding, and their practical use requires determining the appropriate number of particles and strategies to manage particle impoverishment. To overcome these challenges, many algorithms use partially linear approximations, such as Rao-Blackwellised Particle Filters. Quantum computers have the potential to address these challenges by propagating and resampling particles across a larger state space, and future specialised quantum algorithms may be able to directly address non-linear estimation problems. 
However, standardisation is necessary to enable targeted research and comparison of various qubit types and parameters, such as the number of qubits, correlation, and coherence time, before these advancements can be feasible.

\begin{figure*}[htp]
    \centering
    \includegraphics[width=0.8\textwidth]{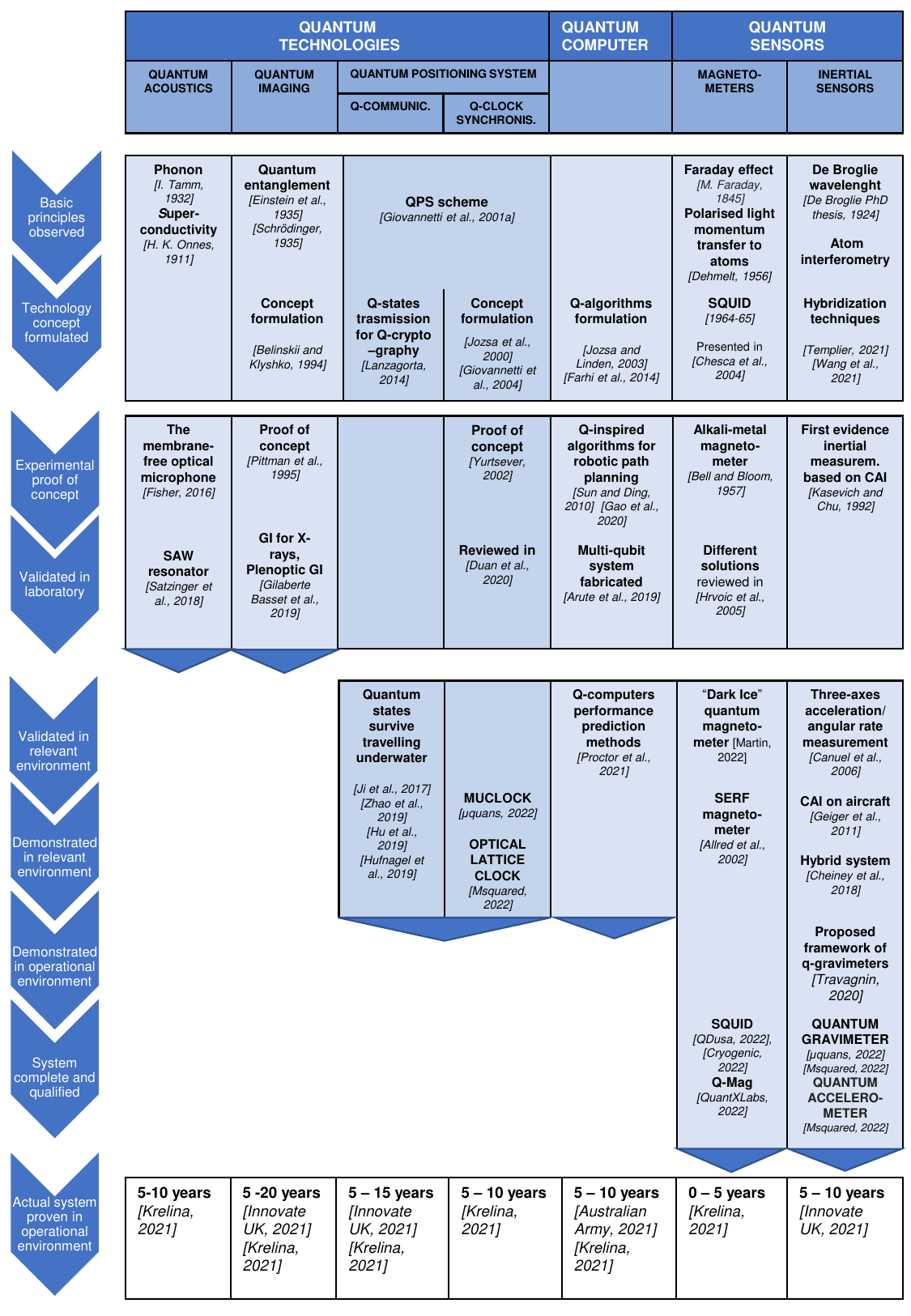}
    \caption{Quantum technology \gls{TRL}~\cite{trl-level-doc} and expected maturity timeline.}
    \label{fig:quantum-trl}
\end{figure*}

\subsubsection{Standardisation and Regulation in Quantum-based Navigation Systems}
It is expected that future quantum-based navigation systems will likely be realised composing the different technologies and the sensors discussed in this work, similarly to what is done today with classical systems.
The level of navigational improvement promised by quantum technology is such that each new quantum sensor might in itself be able to bring substantial improvements to underwater navigation, while it is also possible to see quantum sensors used together with classical instruments in hybrid configurations, especially during periods of technology transitions.
For example, quantum-\gls{INS} might be used together with traditional \glspl{DVL}, classical communications might be enhanced using quantum-level security or classical systems might be used as fallback for those cases where quantum-sensors are not able to provide the desired level of robustness.
It is also possible to imagine quantum-based navigation systems obtained composing a number of quantum sensors that can be used during different phases of a mission or for different mission requirements.
For example, \gls{QPS} can be used when vehicles are at the surface and to initialise quantum-\gls{INS} that can then be used during underwater missions.

One classical distinction with existing navigation systems is based on their need to use external references (e.g., \gls{LBL} or terrain-based navigation) or not ~\cite{Paull:2014}.
In the former case, the system is typically active, which means that it emits signals to reference itself with respect to external references, hence potentially giving away information to neighbouring nodes.
In the latter case instead, the system is only reliant on proprioceptive information and no information is shared with third parties.
Given the current level of technology maturity (most technology is less than \gls{TRL} 5), it is difficult to predict whether quantum-sensors might change this distinction.
At the same time, some results, such as the ability of quantum-cameras to work at sub-shot noise, or produce ghost images, seem to suggest that future quantum sensors might be able to thin this barrier, making possible for active systems to substantially reduce the amount of information distributed to potentially malicious neighbouring listeners.\\

More broadly, as quantum technology evolves, standardisation and regulation is expected to play an important role to deliver confidence to users.
While specific technology will certainly be impacted, for example quantum enhanced imaging could have applications for medical imaging devices only once the regulatory approval is acquired, this is expected to have a more limited impact for maritime navigation, where each sensor manufacturer might be free to define the limits and requirements of their technology.
Technology niches that might require broader international adoption, e.g., \gls{QPS}, is however expected to rely on international standards and need explicit approval. 
This could delay the application of quantum technology further if not actioned promptly.

\section{Conclusions} 
\label{sec:conclusions}
This paper presented the state-of-the art in quantum technology with a focus on the ones that could have applications for maritime navigation.
A range of technologies were discussed with different readiness levels: from quantum computers that could be applied to compute high-accuracy navigational solutions that are today considered too complex (e.g., Monte Carlo methods), to inertial sensors and magnetometers that are starting to become of a size compatible with the dimensions of today's marine autonomous vehicles, to quantum clocks and positioning systems that are also finding their way toward commercialisation.
Less mature technologies include quantum imaging and, even more so, quantum acoustics.
For all these technologies the paper discussed their maturity levels and their main advantages and limitations with respect to classical methods. Quantum technologies might hold the key to solve a number of existing challenges in robotic navigation, and the paper includes indications, based on the authors' findings, on what might be promising technologies that could have an important impact for maritime autonomous navigation.


\appendices
\section*{APPENDICES}
\label{sec:appendix}
This appendix is divided into two sections: Quantum sensing and Quantum entanglement state preparation. The first describes the two main quantum technologies used for inertial navigation: atomic sensors and \gls{SQUID} sensors; and the principles underlying their operation. In the second one the main techniques used for the preparation of entangled states are described.

\section{Quantum sensing}
\label{sec:quantum-sensing}
Inertial navigation requires sensors able to measure physical quantities that include magnetic and electric fields, rotations and translations, gravity, etc. with the greatest possible accuracy. 
Quantum mechanics is makes it possible to use new measurement techniques that are able to surpass classical limits in terms of sensitivity and precision.
This set of new techniques is named \textit{quantum sensing} \cite{Qsensing2017}, and it refers to the use of a quantum features to measure physical quantities.
Among the various types of quantum sensors, this appendix will mainly focus on those that can be used for vehicle navigation: \textbf{atomic sensors} for inertial motion, electromagnetic and gravitational field measurements, and \textbf{superconducting circuits} for magnetic field measurements.\\
It is possible to define quantum sensors through a list of four fundamental characteristics.
They include three of the original requirements defined by~\cite{divincenzo2000} for the physical implementation of quantum computation:

\begin{itemize}
    \item \textit{The quantum system has discrete, resolvable energy levels. For simplicity we will assume that it has two energy levels, the lower energy one, called ground state, indicated with  $|0\rangle$ and an upper energy state, indicated with $|1\rangle$}\footnote{Attention to two-level systems is not a severe restriction, for details see \cite{QLimits2010}.} .\\
    In contrast to classical physics in which the energy of a system can assume any value in the continuum, in quantum mechanics a system can only assume discrete values of energy, called energy levels. 
    In the following we will mainly describe the energy levels of electrons in an atom or molecule, where the transition between one energy level and the next can occur through the absorption or emission of a photon (to rise and fall in level respectively) of frequency $\nu$. 
    For the transition to take place, the energy difference between the two levels must be equal to $E = h\nu = hc/\lambda$,
    where $h=6.62607 \times 10^{-34} J \cdot Hz^{-1}$ is the Planck constant, $c$ the speed of light and $\lambda$ photon's wavelength \cite{nielsen_chuang_2010}
    \item \textit{It must be possible to initialize the quantum system into a well-known state and to read out its state}.\\
    This state could be, for example, a spin orientation with respect to an external magnetic field or an energy state related to the internal structure of the atoms. 
    \item \textit{The state quantum system can be coherently manipulated typically by time-dependent fields}.
    \item \textit{The quantum system evolves as a result of interaction with a physical quantity $V(t)$, such as an electric or magnetic field.}
    At the end of the process the altered state of the quantum system is measured and the effects of the presence of the field are deduced.
\end{itemize}

\subsection{Atomic sensors}
\label{sec:atomic-sensors}
Atomic sensors can be divided in two classes \cite{atomicsensors}: those based on room-temperature atoms confined in an airtight box about $2 mm$ in size (i.e., vapor cell), mostly used for magnetometry and electrometry, and those based on laser-cooled atoms confined by optical and magnetic fields, at the base of the best-performing gravimeters, gyroscopes, and clocks. 
The rest of the section explains the physical principles behind atomic sensors and goes into details on how atomic magnetometers and atomic interferometers work (see also Section~\ref{sec:quantum-magnetometers}). 
In both cases atoms interact with an external field. 
This interaction is mathematically dominated by a quantity called $H_{int}$ which represents the energy of the interaction. 
In general 
\begin{equation}
    H_{int}=\alpha F \equiv \hbar \omega_0
    \label{eq:Hint}
\end{equation}
where $F$ is the field to be measured, $\alpha$ the property of the atom affected by the field, $\omega_0$ the frequency at which the state evolves and $\hbar=h/{2\pi}$ the \textit{reduced Plank constant}. For example for inertial sensors $H_{int}=\Vec{L}\cdot\Vec{\Omega}$, so $\alpha=\Vec{L}$ is the angular momentum of the atom and $F=\Vec{\Omega}$ the rotation rate, or in the case of magnetometers $H_{int}=\Vec{\mu}\cdot\Vec{B}$, hence $\alpha=\Vec{\mu}$ is the magnetic moment of the atom and $F=\Vec{B}$ the magnetic field.

\subsubsection*{Atomic magnetometers}
\label{sec:atomic-magnetometer}
Atomic magnetometers, also called \gls{OAM}, rely on the interaction between atoms and light. More precisely, via analysis of property of light (such as the intensity or polarisation) transmitted through a medium made of atoms prepared in a particular state, it is possible to verify the presence and quantify the intensity of an external magnetic field. The operating principle of an atomic magnetometer is shown in Figure~\ref{fig:OAM}. \\
Initially we confine an ensemble of alkali metal atoms (K, Rb and Cs) in a cell. In the natural state, the orientation of the total angular momentum of an atomic ensemble is undetermined, that is every single orientation is equally probable. We specify that, in quantum mechanics, the total angular momentum of a system, denoted by $\Vec{J}$ is the combination of two types of angular momentum: orbital angular momentum and spin angular momentum. The first refers to a rotation of the system with respect to a chosen centre of rotation, the second is an \textit{intrinsic property} of the system. Although classically one can visualise the spin as the angular momentum about the object's centre of mass, quantum interpretation is somewhat less intuitive and is beyond the scope of this paper (see 
\cite{Foot} for a more detailed discussion).

\begin{figure*}[htp]
\centering
\includegraphics[width=\textwidth]{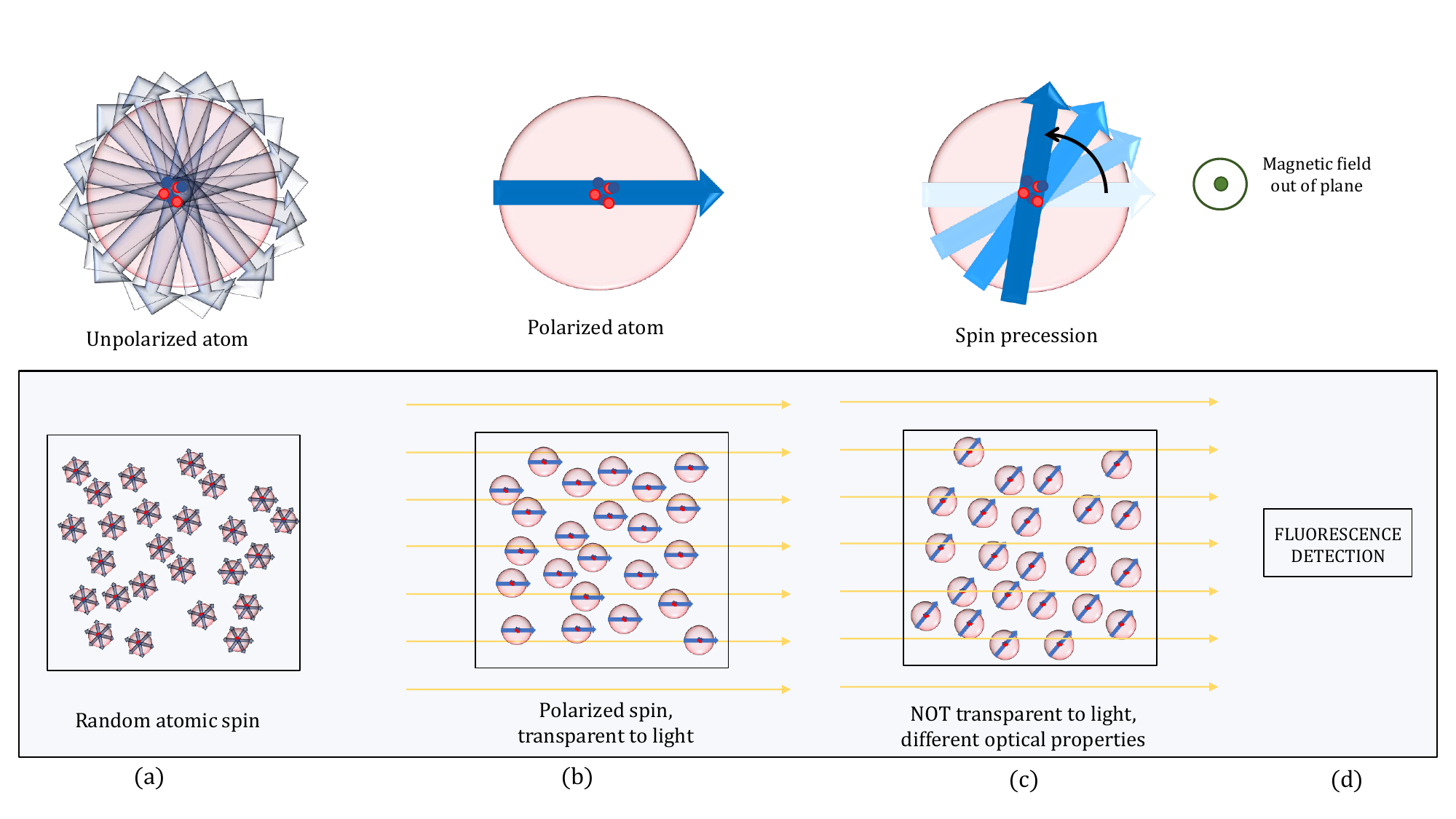}
\caption{Operating principle of an atomic magnetometer. (a) Cloud of unpolarised atom with random
atomic spin, (b) application of circularly polarised light to obtain polarised spins (transparent to light), (c) Spin precession due to an external magnetic field,
and (d) Detection of optical rotation angle.}
\label{fig:OAM}
\end{figure*}

Subsequently we reorient the total angular momentum of atoms by illuminating them with a circularly polarised optical field resonant with an optical (electronic) transition in the atoms. This means that the oscillation frequency of the electromagnetic wave is proportional to the  energy difference between two discrete energy levels of the internal structure of the atoms. In this way, the atom gains angular momentum and eventually becomes polarised along the direction of propagation of the optical field. As a result, when the atomic ensemble is fully polarised it cannot take any more angular momentum and the entire beam of light can be transmitted. In other words the system  becomes transparent to light.\\
The total angular momentum $\Vec{J}$ has a magnetic moment associated $\Vec{\mu}$, linked by a proportionality term $\Vec{\mu}= \gamma\Vec{J}$ called \textit{gyromagnetic constant}. This means that, according to the physical phenomenon called \textit{Larmor precession}, if we now apply an external magnetic field to the polarised system of atoms, this will begin to precess around the magnetic field. Since the spin orientation varies due to precession, the optical properties of the system will change again. Not being perfectly aligned to the light beam direction anymore, the system will no longer be transparent and this will cause a change in power transmitted through the cell or (equivalently) the fluorescence from the atoms.

\subsubsection*{Atom interferometry} 
\label{sec: Atom interferometry}
Atom interferometry is the technique underlying most of the quantum inertial sensors, like accelerometers, gyroscopes, and gravimeters, whose purpose is to extract information from wave interference. To explain how it works, it may be useful to briefly describe the concept of interference, how it is used in optical interferometers and how they differ from atomic ones. \\

Interference occurs when two waves combine to create a new wave with a bigger, smaller, or same amplitude. Positive or negative interference can result from two waves that are coherent or correlated with one another, such as when they have the same frequency or originate from the same source. This principle is used in interferometry to combine waves so that the intensity pattern that results from their interaction reveals information about the phase difference between the waves: waves that are in phase will experience constructive interference while waves that are out of phase will experience destructive interference. An intermediate intensity pattern, present in waves that are neither fully in phase nor fully out of phase, can be utilised to calculate the relative phase difference.

Interferometers can be divided in \textit{common-path} and \textit{double-path interferometers}. In the first class the two beams, called \textit{reference beam} and \textit{sample beam}, travel along the same path, but they may travel along opposite directions, or they may travel along the same direction but with the same or different polarization. Instead, in the second class the reference beam and sample beam travel along divergent paths. A typical exemple is the Mach - Zehnder interferometer (\gls{MZI}), where a laser beam incident from the left is coherently divided by a beam splitter, reflected by mirrors, and recombined by a second beam splitter, (see Figure~\ref{fig:MZ_interferometer:1}-(right)). A photodetector monitors the optical power after the second beam splitter, providing a sensitive measure of shifts in the optical phase difference, $\Delta\phi$ between the interferometer arms. 

The principle of the atomic interferometer, as well as the optical one, is to split and recombine atom matter waves in order to yield interference patterns. Splitting and recombination are done by exploiting the so-called simulated Raman transition. It is a process that involves two photons: one called pump photon, with energy $\omega_p$, and one called Stokes photon, with energy $\omega_s$. The first one is absorbed by the atom (or molecule) bringing it into an excited energetic state, and the second is emitted by the atom causing a de-excitation. If the difference $\omega_p-\omega_s$ corresponds to the energy difference between the two energy levels considered, the energy transition occurs. A scheme of the process is shown in Figure~\ref{fig:MZ_interferometer:1}-(left). In an atomic interferometer, atoms interact not with single photons, but with laser beams - laser 1 and laser 2 - made by photons with energy $\omega_p$ and $\omega_s$, respectively.  In this way the beam is divided in two, one with atoms where the transition took place and the other with the remaining atoms. After the splitting, beams are deflected and recombined by atom optics, which are either laser pulses or mechanical gratings. If there are external forces present during this sequence then there will be a phase shift introduced due to the rotation or linear accelerations. By controlling the intensity and duration of the laser pulses one can prepare the state of an atom in a superposition of the two coupled states with controlled weights. In fact, the percentage of atoms that will undergo the transition is dominated by the intensity and duration of the laser pulses. If these are designed to create the so-called $\pi/2$-pulse the atoms will be in a quantum superposition state for each atom such that it has equal probability to be in states $|0\rangle$ and $|1\rangle$. This is equivalent to a $50-50$ atomic beam splitter. A pulse of this type is called $\pi/2$ because it induces a rotation of $\pi/2$ in the Bloch sphere, bringing the atom into an intermediate state between $|0\rangle$ (pointing along $+z$) and $|1\rangle$ (pointing along $-z$). After a time interval $T$ the atoms are intercepted by optical reflectors, made with a twice longer pulse, a $\pi$-pulse, which swaps the two states, acting as a mirror for the matter wave. Finally, after another period $T$, a second beam splitter ($\pi/2$-pulse) combines the two paths and the fluorescence signals from the atoms 
is read out by the detection system. The total time elapsed by the atoms in the interferometer is $2T$ and is called the interrogation time.

\begin{figure*}[htp]
  \centering
  \includegraphics[width=\textwidth]{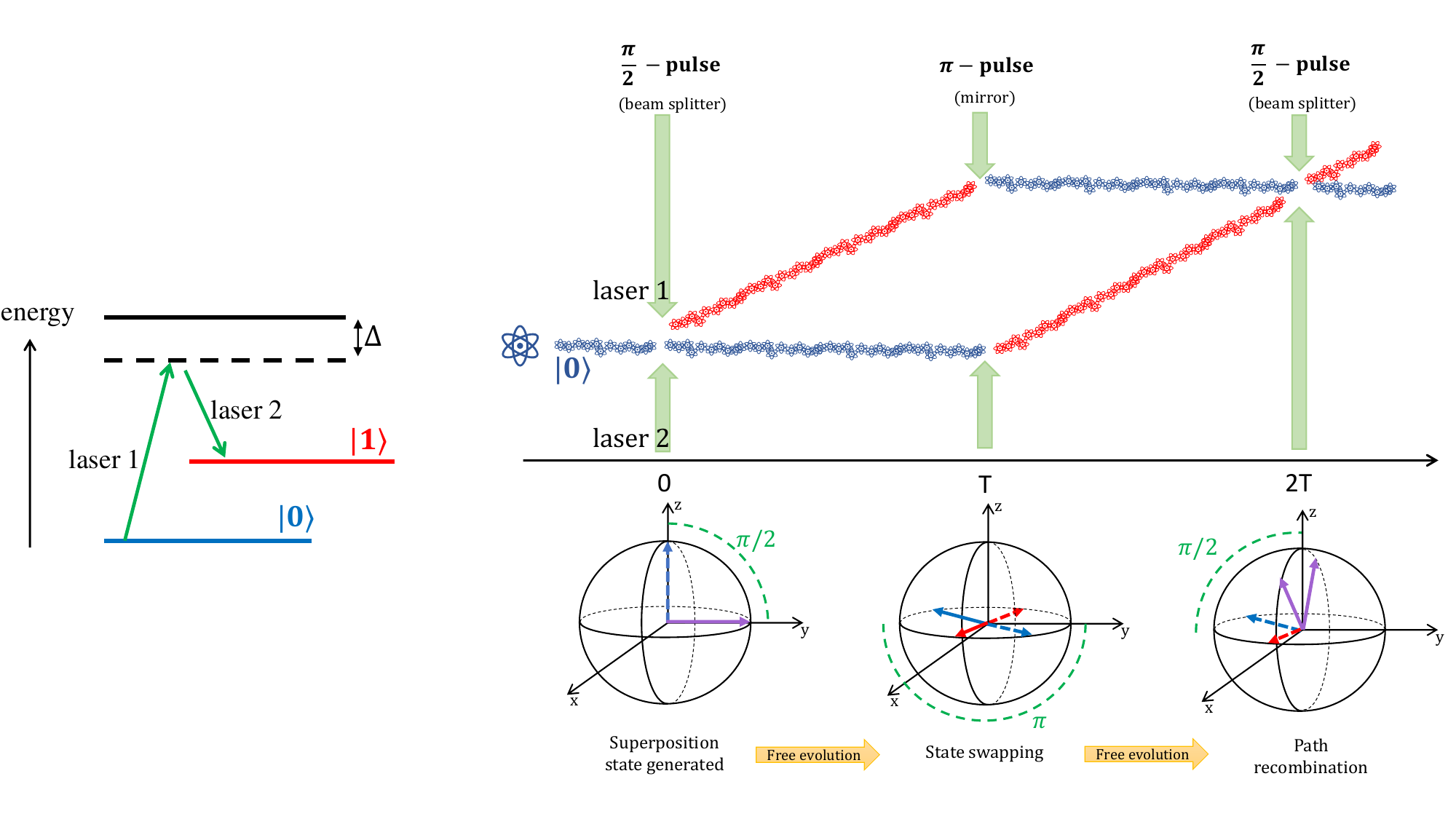}
  \captionof{figure}{On the left: Three-level atom coupled to two counter-propagating laser beams. The atom is subject to a two-photon process called stimulated Raman transition. It absorbs a photon from laser 1 - with energy $\omega_p$ -, reaching a higher energy level (dashed line), and then the  emission of a photon in the mode of laser 2 - with energy $\omega_s$ -is stimulated; the energy of the atom decreases again and, if $\omega_p-\omega_s$ corresponds to the energy difference between $|0\rangle$ and $|1\rangle$, the atom ends up in the state $|1\rangle$. On the right: atomic Mach - Zehnder interferometer scheme. Atoms are initially in state $|0\rangle$ and then through a sequence of $\pi/2 - \pi - \pi/2$ pulses they are first splitted into two parts, inverted and finally recombined in order to observe the interference signal.}
  \label{fig:MZ_interferometer:1}
\end{figure*}

Measuring the two populations $N_1$ and $N_2$ (the fraction of atoms emerging from the interferometer in states $|0\rangle$ and $|1\rangle$, respectively) we derive the transition probability $P = N_1/(N_1+N_2)$ which is proportional to the phase difference $\Delta \phi$ between the two arms of the interferometer.
Since the phase difference is proportional to the external inertial forces to which the atoms are subjected (see equations. (\ref{eq:phi_acc}), (\ref{eq:phi_rot})), this measurement is used to estimate accelerations, as in the case of accelerometers, gravimeters etc., and rotations as in the case of gyroscopes.
In the first case the Raman pulses are applied along and against the gravity vector and, over a time $t = 2T$, the phase shift accumulated is proportional to the acceleration $g$
\begin{equation}\label{eq:phi_acc}
    \Delta\phi = -(k_1+k_2)gt^2 = -k_{eff}gt^2
\end{equation}
where $k_1$ and $k_2$ are laser 1 and laser 2 wavevectors, respectively.\\
In the second case the atoms have an initial velocity orthogonal to the
direction along which the stimulated Raman pulses are applied and in this arrangement a rotation $\Omega$ causes a phase shift
\begin{equation}\label{eq:phi_rot}
     \Delta\phi = 2\pi\Omega k_{eff}vT^2
\end{equation}
with $v$ being atom velocity.

\subsection{Superconducting circuits} 
\label{sec:squids}
Aside from \gls{OAM}, the current state of the art technology in weak-field magnetometry includes \gls{SQUID}. \glspl{SQUID} are extremely sensitive magnetometers capable of detecting very weak magnetic fields. 
There are three key quantum mechanical effects crucial to the operation of a \gls{SQUID}:

\begin{itemize}
    \item Superconductivity\\
    Superconductivity is a physical phenomenon whereby the resistance of a material spontaneously drops to zero if it is cooled below a certain temperature, called \textit{critical temperature} (different for each material). 
    With reference to Figure~\ref{fig:squid}, the resistivity of non-superconducting materials (e.g., gold, silver,...) decreases with temperature until a defined material-dependent finite value, after which point their resistivity remains constant.
    Superconducting materials (e.g., aluminium, lead, titanium...) instead are characterised by a critical temperature $T_c$ below which their measurable resistivity goes to zero, and are hence able to conduct electricity with extremely high efficiency and with almost no energy loss. The critical temperature is different for each material and can range from a few $mK$ to tens of $K$.
    It is easy to see how materials with zero resistance can be useful, but it is important to highlight that the required critical temperatures can be very difficult and expensive to obtain.
   
    \item Quantum tunnelling\\
    Quantum tunnelling is a quantum mechanical phenomenon whereby the wavefunction of a particle, which represents its state, has some finite probability to propagate through a potential energy barrier, even if its energy E is below the height of the barrier V. Imagine a particle with energy $E$ confined in a potential well (the region surrounding a local minimum of potential energy) with a maximum value $V$. In classical physics, if $E<V$ the particle cannot cross the energetic barrier and remains in the well forever. In quantum mechanics the particle has a non-zero probability to tunnel through the potential barrier and emerge with the same energy $E$.\\
    The tunnel effect is the basis of the functioning of the Josephson junction, one of the fundamental components of a \gls{SQUID}. This is composed of two superconducting strips separated by an insulating layer.
    
    \item Magnetic flux quantisation\\
    The magnetic flux, represented by the symbol $\Phi$, is defined as $\Phi = B \cdot S$, where $B$ is the magnetic field and $S$ the area crossed by the field. 
    If the crossed area is surrounded by a superconducting material (superconducting loop or a hole in a bulk superconductor), the magnetic flux is quantised, and the field can assume multiple integer values of the elementary quantity:
    
    \begin{equation*}
        \Phi_0 = \frac{h}{2e} = 2.067\ 833\ 636\times10^{-15} \ \text{Wb}
    \end{equation*}
    
    where $h$ is the Plank constant and $e = 1.602\ 176\ 634\times10^{-19} \ C$ the elementary charge. 
    The discrete and indivisible quantity of magnetic flux is a physical constant, as it is independent of the underlying material as long as it is a superconductor.
    
\end{itemize}

\begin{figure}[htp]
\centering
\includegraphics[width=0.35\textwidth]{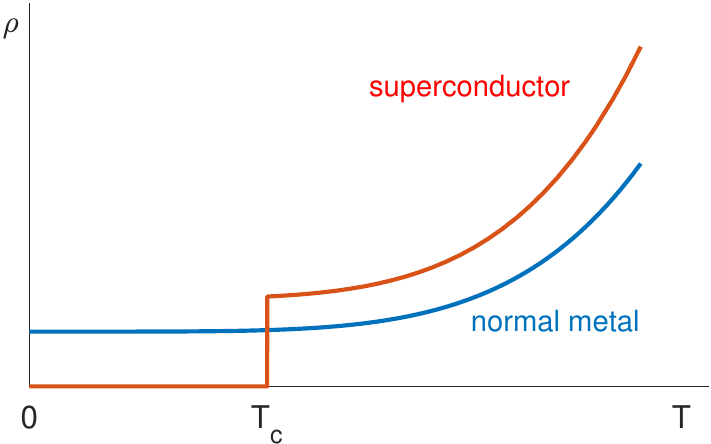}
\centering
\includegraphics[width=0.5\textwidth]{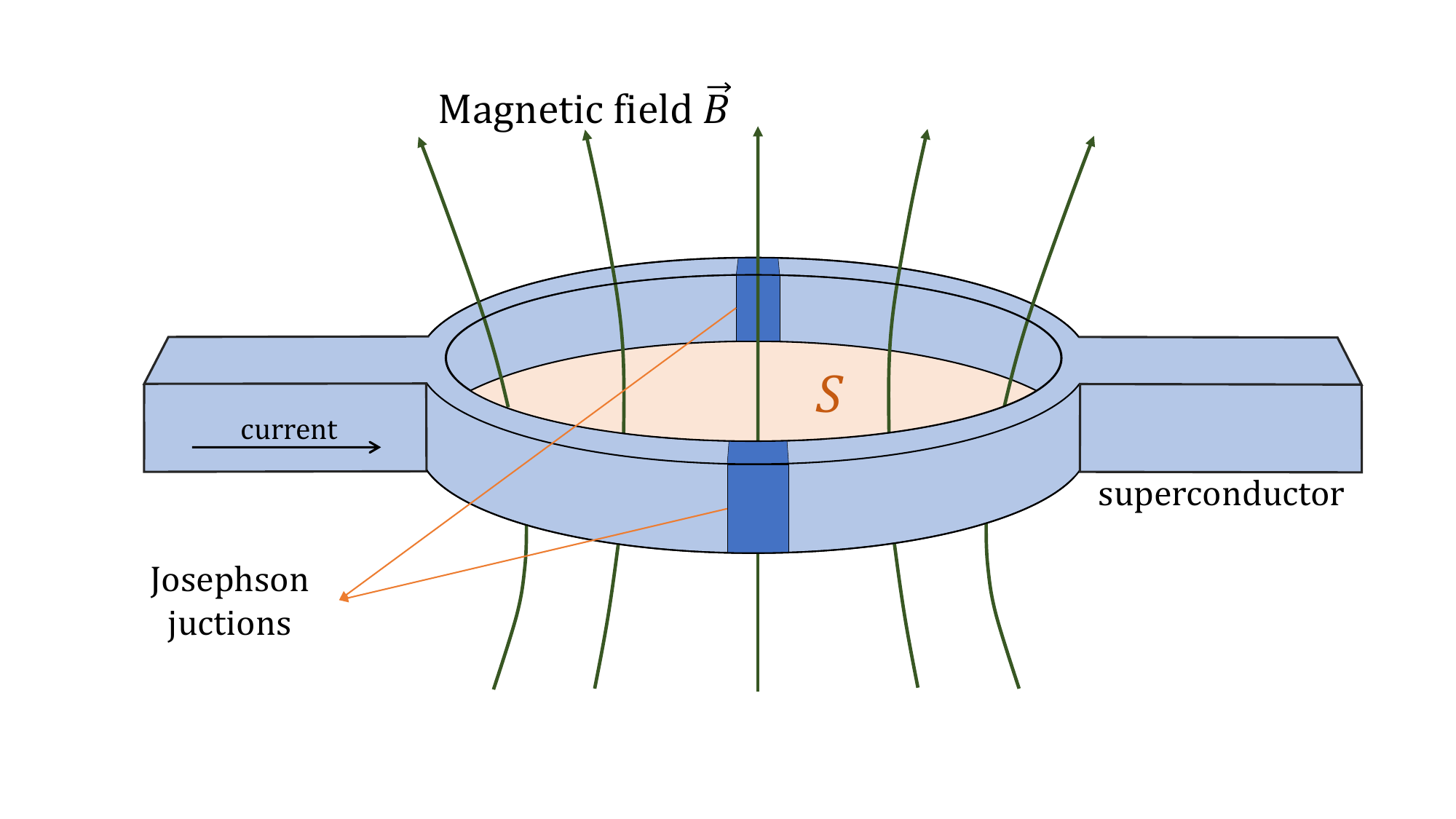}%
    \caption{Left: Typical trend of the resistance $\rho$ as a function of temperature $T$, for non-superconducting and superconducting materials. Right: Sketch of a SQUID.}
    \label{fig:squid}
\end{figure}


A \gls{SQUID} device, shown in the Figure~\ref{fig:squid} is made up of a superconducting ring containing one or more Josephson junctions. 
To date \glspl{SQUID} are among the best commercial quantum magnetometric sensors but have the big disadvantage that they require cryogenesis.

\section{Preparation of quantum entangled states} 
\label{sec:quantum-entangled-states}
At present, there are various methods for preparing entangled states, including \gls{SPDC} in nonlinear crystal \cite{mandel_wolf_1995}, the ion trap method  and the \gls{c-QED} 
\cite{nielsen_chuang_2010}, of which we will give a brief description below. 
Among these, the \gls{SPDC} method is the most versatile and efficient one as it does not require cooling of ion motion such as ion traps, and is more resistant to decoherence phenomena than \gls{c-QED} \cite{QPSsurvey}.\\

\subsection{Spontaneous Parametric Down Conversion - SPDC}
\label{sec:spdc}
 In a typical \gls{SPDC} setup,  
 a laser beam is directed towards a non-linear optical medium (e.g., a $\beta$-Barium-Borate (BBO) crystal). 
 Thanks to this interaction, with a certain probability, an incident photon, called \textit{pump}, decays into a pair of related photons, called \textit{signal} and \textit{idler}. 
 In optics, the non-linearity of a medium refers to the electric polarisation $P$, a quantity that describes how a material responds to an applied electric field $E$. In \gls{SPDC} process a crystal with a second order non-linearity is used, which means that the electric polarisation contains, in addition to a term proportional to $E$, also one proportional to $E^2$. This process satisfies the conservation of energy and conservation of momentum laws
 
 \begin{align}
    \omega_p = \omega_s+\omega_i \\
    \textbf{k}_p = \textbf{k}_s+\textbf{k}_i    
 \end{align}

where $\omega_p$, $\omega_i$ and $\omega_s$ are frequency of pump light, idle light and signal light; $\textbf{k}_p$, $\textbf{k}_i$ and $\textbf{k}_s$ the corresponding light wave vectors. Figure~\ref{fig:SPDC} shows a scheme of the \gls{SPDC} process.

\begin{figure}[htp]
\centering
  	\includegraphics[width=0.5\textwidth]{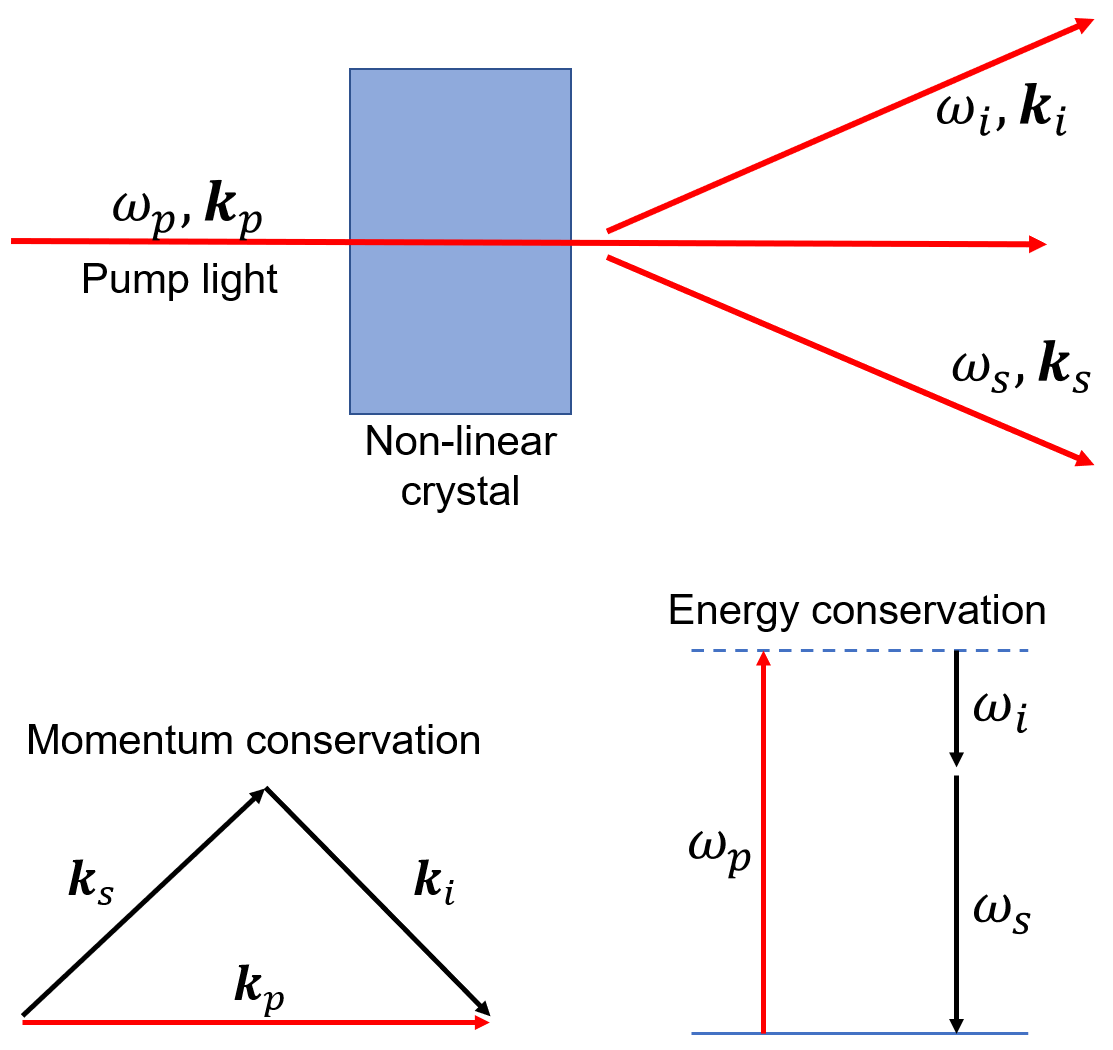}
  	\caption{SPCD process. A photon (or a beam of photons, called \textit{pump}) passes through a non-linear crystal. With a certain probability, the pump photon can decay into a pair of lower energy photons, called \textit{signal} and \textit{idler}, satisfying the conservation of momentum and energy laws.}
     \label{fig:SPDC}
\end{figure}



The entangled photon pairs generated by \gls{SPDC} can be entangled in different properties, including their momentum, time-energy \cite{EnergyTimeEntanglement},  polarization \cite{PolarizationEntanglement} (critical for quantum computation) and space \cite{WALBORN201087} (mainly used for quantum imaging).

\subsection{Ion trap}
\label{sec:ion-trap}
Another method to achieve entangled states of two or even more atoms is to use ion traps. An ion trap is a device capable of capturing charged particles - known as ions - with the help of electric and magnetic fields. 
Paul trap is the most used ion trap type for studies of quantum state manipulation, which forms a potential energy via a combination of static and oscillating electric fields. This combination is necessary because it is impossible to confine an ion using a purely electrostatic field due the Earnshaw’s theorem. The trap consists of two hyperbolic metal electrods facing each other, and an hyperbolic ring electrod between the first two, as shown in Figure~\ref{fig:IONTRAP}. The trapped ion experiences a potential energy similar to that of a ball on a saddle-shaped surface. Rotation of the surface about a vertical axis, at a suitable speed, prevents the ball rolling off the sides of the saddle and gives stable confinement.

\begin{figure*}[htp]
\centering
\includegraphics[width=0.8\textwidth]{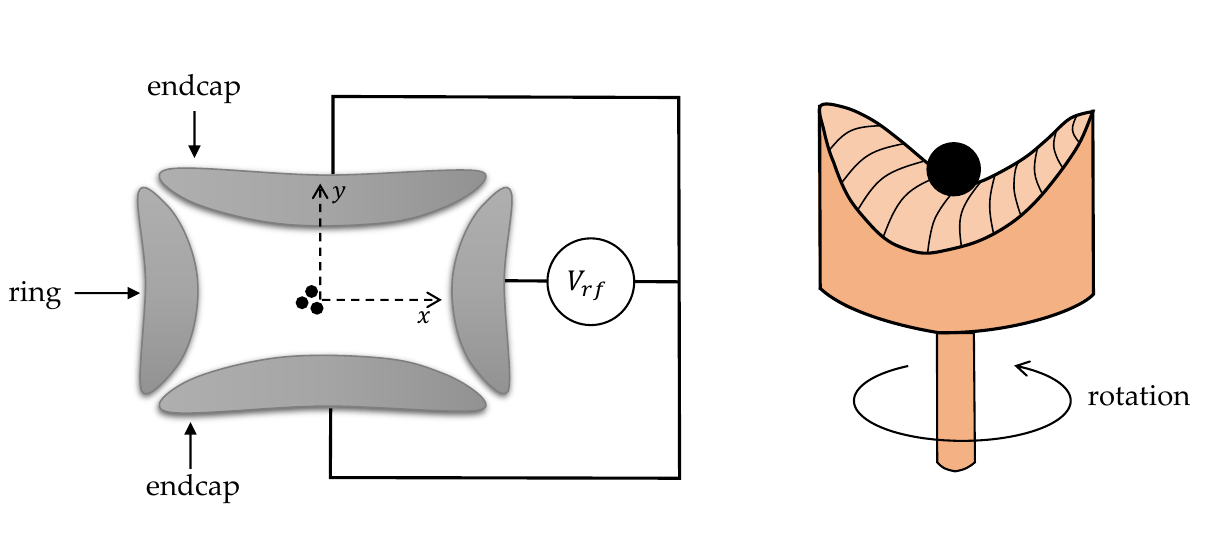}%
\caption{On the left: Paul Ion Trap scheme. A \textit{rf} (radio-frequency) voltage $V_{rf}=cos(\omega_{rf}t)$ is applied between the endcap and ring electrodes. On the right: an intuitive analogy between the potential energy experienced by trapped ions and that of a ball on a rotating saddle-shaped surface. Adapted from ~\cite{Deng2007}.}
\label{fig:IONTRAP}
\end{figure*}

Lasers make it simple and precise to manipulate trapped ions. Trough this interaction is possible to prepare and control quantum states relating trapped ion's internal states  with their motion (i.e., normal modes). Furthermore by using laser cooling, the trapped ions can get close to a temperature below 1K and create an equilibrium crystal structure where their attraction to one another balances the potential for trapping. This method has two main advantages: first, because the ions are trapped in a high vacuum environment, they are almost in isolation without interference, and have a long time of decoherence; the second the preparation of initial state and the measurement of quantum state have extremely high fidelity and efficiency. \\

\subsection{Cavity Quantum Electrodynamics (c-QED)}
\label{sec:cQED}
The study of the preparation of entangled states by cavity quantum electrodynamics is gradually established with the development of cold atom technology and photoelectric testing technology. 
The two main experimental components of a \gls{c-QED} system are the electromagnetic cavity of very high quality factor $Q$ (ratio between the stored electromagnetic energy of the cavity and the power losses) and a single atom placed inside it. 
The idea is to store quantum information in the atomic energy state after trapping it in the cavity and making it interact with the field inside.
The interaction can vary depending on the geometry of the cavity, the interaction time and the intensity of the field and various entangled
states of atoms and light fields can be prepared (\cite{GUO1997}, \cite{guo2000}, \cite{Rfifi2016}). 
The main disadvantage of this scheme is that, by using this type of interaction as a memory mechanism, any disturbance in the environment (e.g., decoherence of the cavity or dissipation in the circuits) risks breaking the correlation (disentanglement) frustrating the process \cite{Zueco_2010}.

\bibliographystyle{IEEEtran}
\bibliography{andrea,riccardo,olga}


 




\vfill

\end{document}